\documentclass[conference]{IEEEtran}
\IEEEoverridecommandlockouts
\usepackage{cite}
\usepackage{amsmath,amssymb,amsfonts}
\usepackage{algorithmic}
\usepackage{graphicx}
\usepackage{textcomp}
\usepackage{xcolor}
\def\BibTeX{{\rm B\kern-.05em{\sc i\kern-.025em b}\kern-.08em
    T\kern-.1667em\lower.7ex\hbox{E}\kern-.125emX}}

\usepackage{wrapfig}
\usepackage{mathtools}
\usepackage{subcaption}
\usepackage{bm}
\usepackage{multirow}
\usepackage{booktabs}

\graphicspath{{figures/}{figures/}}
\def\Agents{\mathcal{I}}
\def\ActionSet{\mathcal{A}}
\def\sal#1{\bm{#1}^{\mathrm{sal}}}
\def\env{\mathcal{E}}

\usepackage{xspace}

\def\ignore#1{\iffalse #1 \fi\xspace}

\def\ignore#1{\xspace}

\usepackage[normalem]{ulem}

\begin{document}

\title{Interpretability for Conditional Coordinated Behavior in Multi-Agent Reinforcement Learning}

\author{\IEEEauthorblockN{Yoshinari Motokawa}
\IEEEauthorblockA{\textit{Dept. of Computer Science} \\
\textit{Waseda~University, Tokyo, Japan}\\
y.motokawa@isl.cs.waseda.ac.jp}
\and
\IEEEauthorblockN{Toshiharu Sugawara}
\IEEEauthorblockA{\textit{Dept. of Computer Science} \\
\textit{Waseda~University, Tokyo, Japan}\\
sugawara@waseda.jp}
}

% \author{\IEEEauthorblockN{Anonymous Authors}}

\maketitle

\begin{abstract}
We propose a model-free reinforcement learning architecture, called \emph{\textbf{d}istributed \textbf{a}ttentional \textbf{a}ctor \textbf{a}rchitecture \textbf{a}fter condition\textbf{a}l \textbf{a}ttention} (DA6-X), to provide better interpretability of conditional coordinated behaviors. The underlying principle involves reusing the saliency vector, which represents the conditional states of the environment, such as the global position of agents. Hence, agents with DA6-X flexibility built into their policy exhibit superior performance by considering the additional information in the conditional states during the decision-making process. The effectiveness of the proposed method was experimentally evaluated by comparing it with conventional methods in an objects collection game. By visualizing the attention weights from DA6-X, we confirmed that agents successfully learn situation-dependent coordinated behaviors by correctly identifying various conditional states, leading to improved interpretability of agents along with superior performance.
\end{abstract}


\section{Introduction}
\emph{Explainable reinforcement learning} (XRL) has attracted considerable attention in the academic and industrial domains. Several XRL categories exist for the interpretability of agents~\cite{survey2}. For instance, the decision-making process of agents has been interpreted by decomposing reward functions~\cite{650899} or visualizing their saliency maps based on the integrated gradients method~\cite{IntegratedGradients}. The \emph{attention mechanism} developed by \cite{vaswani2017attention}, in addition to the transformer, also plays a critical role in imparting transparency to the decision-making process. Consequently, various neural network models based on the attention mechanism and transformer, such as \emph{vision transformer}~\cite{dosovitskiy2021image} and \emph{decision transformer}~\cite{chen2021decision}, have enabled the successful visualization of the input information that plays a key role in achieving the state of the art performance in computer vision and even \emph{deep reinforcement learning} (DRL).
\par

However, few studies have focused on XRL in multi-agent systems (MAS), although clarification of the black-box coordination/cooperation mechanism is crucial in inducing better productivity and robustness for the entire system. \emph{Multi-actor-attention-critic} (MAAC)~\cite{pmlr-v97-iqbal19a}, incorporating the attention mechanism in the style of MADDPG~\cite{MADDPG}, demonstrated the way agents selectively focus on cooperation among themselves using the attention mechanism. \cite{MAT_DQN,DA3} developed an approach to establish the interpretability of agents' coordinated behaviors by analyzing the individual agent's attention mechanism and empirically clarified that their agents would selectively identify other agents worthy of coordination. Such findings are expected to improve the explainability of learned behaviors of individuals as well as the efficiency of the entire system. Meanwhile, agents in MAS must behave flexibly depending on the conditions and situations that include other agents and their actions. Agents can learn such conditional behaviors using the aforementioned methods; however, these methods are not sufficient to ensure that the interpretability of the learning results will make agents acquire the expected conditional behaviors.
\par

To address this challenge, we proposed \emph{\textbf{d}istributed \textbf{a}ttentional \textbf{a}ctor \textbf{a}rchitecture \textbf{a}fter condition\textbf{a}l \textbf{a}ttention} (DA6-X). The model architecture is based on reusing the \emph{saliency vector}~\cite{DA3} to enhance the interpretability of learned conditional coordinated behaviors in \emph{multi-agent DRL} (MADRL). DA6-X is a sequential framework that combines a module for recognition of the conditions/situations, with any DRL head (corresponding to 'X'), to output the learned behavior. Unlike the previous models~\cite{MAT_DQN,DA3}, DA6-X takes two-layered data: \emph{conditional states} (conditions/situations) and local observations. DA6-X first represents the conditional states using the saliency vector in the former \emph{conditional module} (CM). The saliency vector is then reused along with the local observations to feed to the latter \emph{local transformer encoder}, such that DA6-X is capable of comprehensive learning under various situations. In addition, agents with DA6-X (DA6-X) not only learn condition-dependent actions in a more organized manner but also provide information for understanding the justification and rationale of their actions. By extracting the \emph{attention weights} from the local transformer encoder in DA6-X, attention heatmaps are generated to determine the parts of the input data describing the conditional aspects that affect the intensity of interests in the two-layered data, especially local observation. This information also helps us understand how conditional states affect the agents' behaviors, even when the same local observations are provided.
\par

Our primary contributions are summarized as follows:
\begin{enumerate}
  \item DA6-X, a novel MADRL architecture, was designed for interpreting situation-dependent/conditional coordinated behaviors by reusing the saliency vector.
  \item While providing transparency, DA6-X can be easily integrated with any DRL algorithm (hence, \emph{X} is included in the name) in single-agent or multi-agent DRL systems.
  \item DA6-X agents with various conditional states achieve better performance than several existing algorithms taking the \emph{objects collection game} as an example.
  \item Improved interpretable coordinated behaviors are qualitatively demonstrated via the attention mechanism by hierarchically analyzing the input information.
\end{enumerate}
\par

\section{Related Work}\label{Sec:RelatedWorks}
{\bfseries Attention-based method in XRL:}
In addition to developing research on visual explanations, such as feature-based~\cite{Iyer2018TransparencyAE}, embedding-based~\cite{mnih2015humanlevel}, perturbation-based~\cite{10.1007/978-3-319-10590-1_53}, and gradient-based methods~\cite{IntegratedGradients}, incorporation of the attention mechanism in models is one of the most popular methods in XRL~\cite{8957473}. Recently, \emph{DA3-X}~\cite{DA3} was proposed as an extension of \emph{MAT-DQN}~\cite{MAT_DQN} to demonstrate how decentralized agents build coordination by highlighting the influence of relevant tasks, other agents, and the noise in local observations based on the attention mechanism.
\par

However, these prior studies only represent the influential segments of the agents' inputs in an aggregated manner by focusing on the performance improvement and ignoring the dependencies within the inputs, i.e., they do not determine the segments in the inputs that direct the agent's attention to other parts of segments. In contrast, our proposed method explicitly classifies the input information into conditional segments, such as the global positions of the observing agent and other agents, and baseline segments, such as the local visible region, to identify the parts in the conditional segments that affect the agents' focus of interest within their local observations and result in the development of conditional, situation-dependent, and coordinated behaviors.
\par

{\bfseries Attention Mechanism in MADRL:}
Various MADRL models utilize the attention mechanism. Several studies~\cite{MANet,pmlr-v97-iqbal19a} use the attention mechanism as a centralized communication processor that efficiently handles encoded messages among agents in MADRL. Incorporation of the attention mechanism in MAS is also beneficial for constrained problems~\cite{MACAAC}, such as approximation of underlying behaviors of agents~\cite{Li2020GenerativeAN} and trajectory prediction~\cite{li2020Evolvegraph}. In particular, \cite{MANet} introduced the \emph{multi-focus attention network}, which helps agents attend to important sensory-input information using multiple parallel attentions in a grid-like environment. \cite{zambaldi2018deep} investigated the enhancement of agents' ability to efficiently adapt to complicated environments (Box-World and StarCraft II) that require relational reasoning over structured representations using the attention mechanism. \cite{JointAtt} introduced \emph{joint attention}, which aggregates every other agent's attention map and demonstrates its cost-effectiveness in multi-agent coordination environments.
\par

The early studies on MADRL mainly focused on augmenting the agents' performance using centralized attention in the {\em centralized training with decentralized execution} (CTDE) approach; however, the fully decentralized approach is generally more robust because of less variance in policy updates and is more feasible in realistic domains~\cite{Lyu2021ContrastingCA}. Moreover, the qualitative analysis of coordination using attention heatmaps has not been studied in detail. Our goal is to investigate the specific behavioral analysis of cooperative agents based on decentralized attention heatmaps to clarify how to undertake alternative strategies and coordination structures depending on the conditional states and local observations for better interpretability.
\par

\section{Preliminaries}\label{sec:preliminaries}
{\bfseries Dec-POMDP:}
In this study, the focus is on the \emph{decentralized partially observable Markov decision process} (dec-POMDP)~\cite{POMDP} of $N$ agents, denoted by a tuple $\langle\Agents, \mathcal{S}, \{\ActionSet_i\}, p_T, \{r_i\},\{\Omega_i\}, \mathcal{O}, H\rangle$. $\Agents = \{1, \ldots, N\}$ indicates a set of agents; $\mathcal{S}$ is a finite set of states; and $\mathcal{A}_i$ is a finite set of individual action space of $i\in\Agents$. Suppose $a\in\mathcal{A}$ and $s, s'\in \mathcal{S}$; $p_T(s'|s,a)$ is a transition probability; $r_i(s, a)$ is a reward ($\in\mathbb{R}$) obtained by $i\in\Agents$; $\Omega_i$ is a finite set of observations by $i\in\Agents$; $\mathcal{O}(o|s,a)$ is a transition probability $o\in\Omega$; and $H$ is the time horizon of the process. In dec-POMDP, the agents aim to maximize the discounted cumulative reward $R_i = \sum_{t=0}^H {\gamma^t r_i(s,a)}$ by updating their individual policies $\pi_i$, where $\gamma$ is a discount factor ($0\leq \gamma <1$).
\par

\begin{figure}
  \begin{minipage}[t]{0.49\hsize}
    \centering
    \includegraphics[keepaspectratio, width=\linewidth]{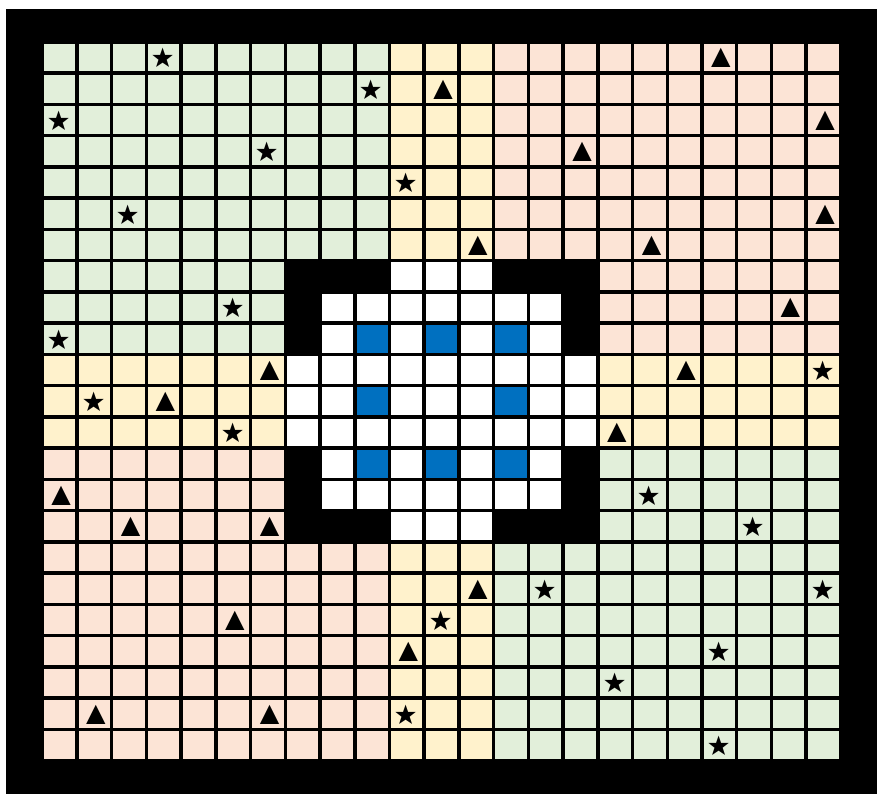}\\
    \subcaption{Object spawn region (green for objects $\bigstar$, red for objects $\blacktriangle$, and beige for both objects $\bigstar$, $\blacktriangle$).}\label{fig:map_environment}
  \end{minipage}
  \begin{minipage}[t]{0.49\hsize}
    \centering
    \includegraphics[keepaspectratio, width=\linewidth]{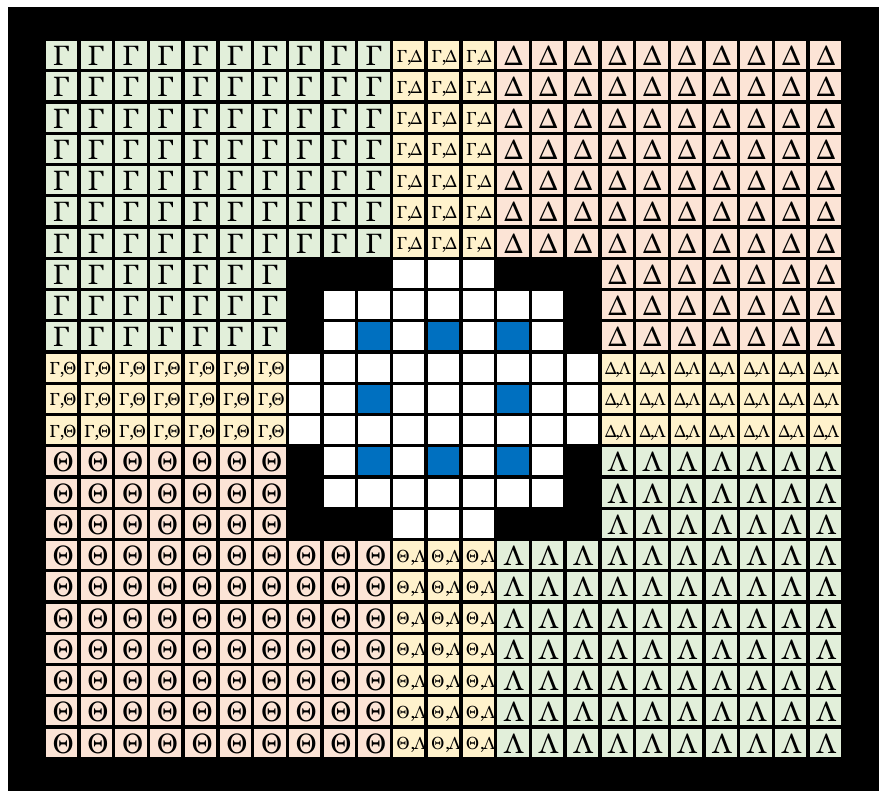}\\
    \subcaption{Task execution area, which is partitioned into four distinct regions labeled as $\Gamma$, $\Delta$, $\Theta$, and $\Lambda$.}\label{fig:map_environment2}
  \end{minipage}
  \caption{A grid environment example.}\label{fig:environment}
\end{figure}

{\bfseries Problem Setting:}
In the object collection game, agents collect as many objects as possible in a grid-like environment of size $G_X\times G_Y$, as shown in Fig.~\ref{fig:environment}. At the beginning of each episode, agents are placed at the initial positions indicated by blue cells in Fig.~\ref{fig:environment} and start exploration. At each step, agents encode the observed local environment in $N_\mathrm{C}$ channels of $R_\mathrm{X}\times R_\mathrm{Y}$ matrices (where $R_\mathrm{X}\times R_\mathrm{Y}$ is the size of the observation matrices $\in \mathbb{R}^{R_\mathrm{X}\times R_\mathrm{Y}}$) and feed their local observation in shape of $N_\mathrm{C} \times R_\mathrm{X}\times R_\mathrm{Y}$ into their own networks. Then, they decide the suitable action $a_i\in \ActionSet_i = \{\mathit{up}, \mathit{down}, \mathit{right}, \mathit{left}\}$, wherein each element describes the movement direction of agents for the next step.
\par

Agent $\forall i$ receives a reward $r_i$ depending on the condition at the next step, i.e., $i$ obtains a positive reward $r_\mathrm{o}>0$ if it moves to the same position as an object; it receives a negative reward $r_\mathrm{c}<0$ if it collides against other agents or walls; otherwise, it receives $r_i = 0$.
\par

{\bfseries Multi-Head Attention:}
The \emph{self-attention mechanism} was introduced by~\cite{vaswani2017attention} to get similarities in sequences as
\begin{equation}\nonumber
    \mathrm{Attention}(Q,K,V) = \mathrm{Softmax}(\frac{Q\cdot K^T}{\sqrt{d}})V,
\end{equation}
where $Q$, $K$, and $V$ denote \emph{query}, \emph{key}, and \emph{value} matrices, respectively, and $d$ is the dimension of the query/key. \emph{Multi-head attention} (MHA) is determined by calculating the self-attention in $h$ parallel attention heads, as shown in equation below:
\begin{equation}
  \begin{split}\nonumber
    \mathrm{MHA}(Q, K, V) &= \mathrm{Concat}(\mathit{head}_1,
    \ldots, \mathit{head}_h)W^O\\
    \mathit{head}_l &= \mathrm{Attention}(Q\cdot W_l^Q, K\cdot W_l^K, V\cdot W_l^V),
  \end{split}
\end{equation}
where $W_l^Q$, $W_l^K$, $W_l^V$, $W^O$ are projected parameter matrices for attention head $\mathit{head}_l$ ($1 \leq l \leq h$). The reader is referred to the original paper~\cite{vaswani2017attention} for more details.
\par

{\bfseries DA3-X:}
DA3-X~\cite{DA3} is a neural network model comprising the attention mechanism, as shown in Fig.~\ref{fig:DA3}. The significance of DA3-X is the interpretability of agents in distributed multi-agent systems. As agents are required to build coordination with other agents in multi-agent systems, it is crucial to clarify how their cooperative behaviors are induced through their black-box decision-making process. A few prior studies worked on the transparency of coordination in centralized multi-agent systems, but mostly they did not assume the distributed system that is more feasible in real-world applications. By using DA3-X as a baseline method, we verified its interpretability as well as its ability to selectively distinguish important segments of observation by analyzing the attention weights in DA3-X. Its versatile network structure where arbitrary reinforcement learning methods can be applied (DA3-DQN, DA3-DDPG, etc.) is also remarkable. Moreover, we verified that it is scale-free with respect to the number of agents as the DA3-X supposes a distributed system and does not depend on models from other agents.
\par

\section{Proposed Method: DA6-X}\label{sec:methodology}
\subsection{Neural Network Architecture}
\begin{figure*}
  \centering
  \begin{minipage}[t]{0.35\hsize}
    \centering
    \includegraphics[keepaspectratio, width=\linewidth]{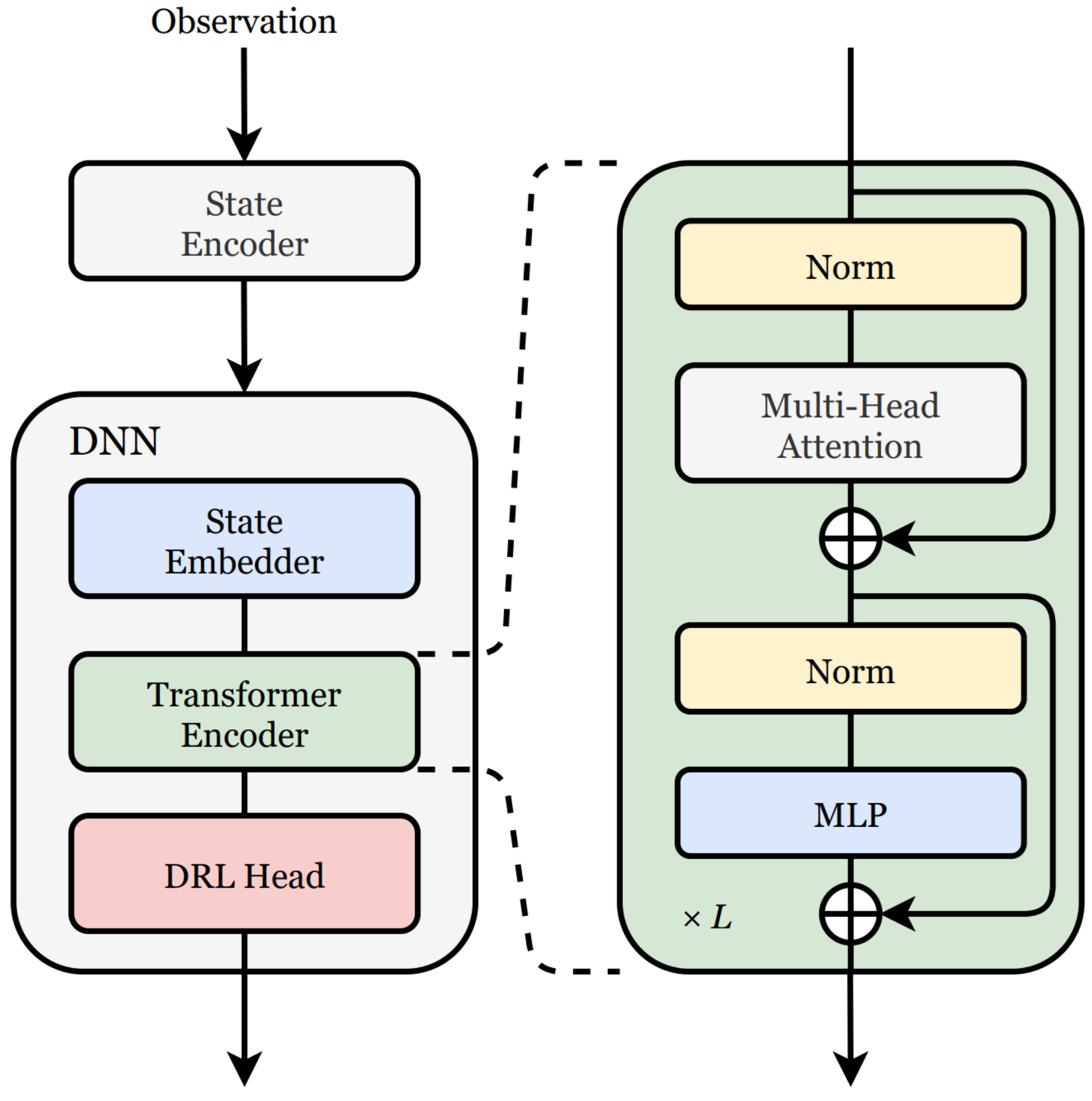}\\
    \subcaption{DA3-X~\cite{DA3}.}\label{fig:DA3}
  \end{minipage}
  \hfil
  \begin{minipage}[t]{0.3\hsize}
    \centering
    \includegraphics[keepaspectratio, width=0.9\linewidth]{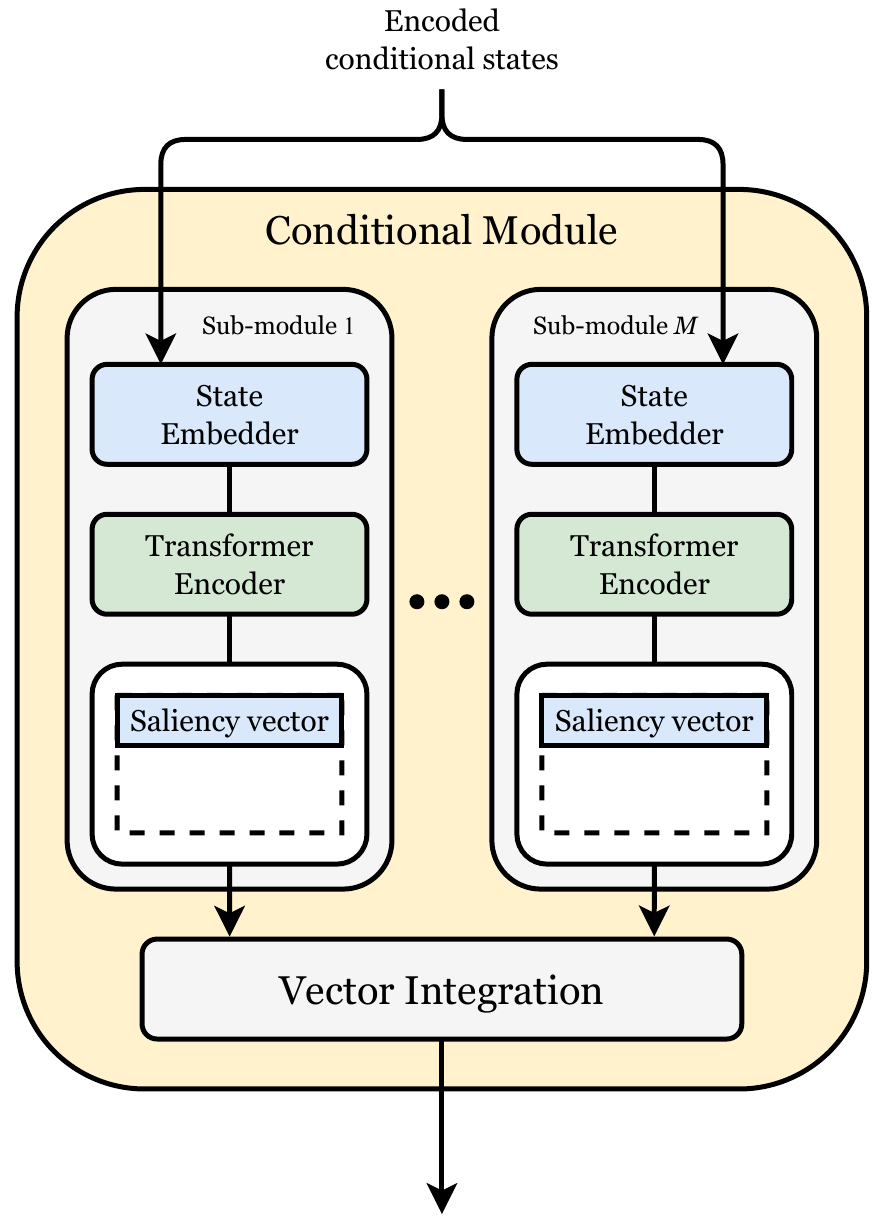}\\
    \subcaption{Conditional Module.}\label{fig:ConditionalModule}
  \end{minipage}
  \hfil
  \begin{minipage}[t]{0.3\hsize}
    \centering
    \includegraphics[keepaspectratio, width=0.85\linewidth]{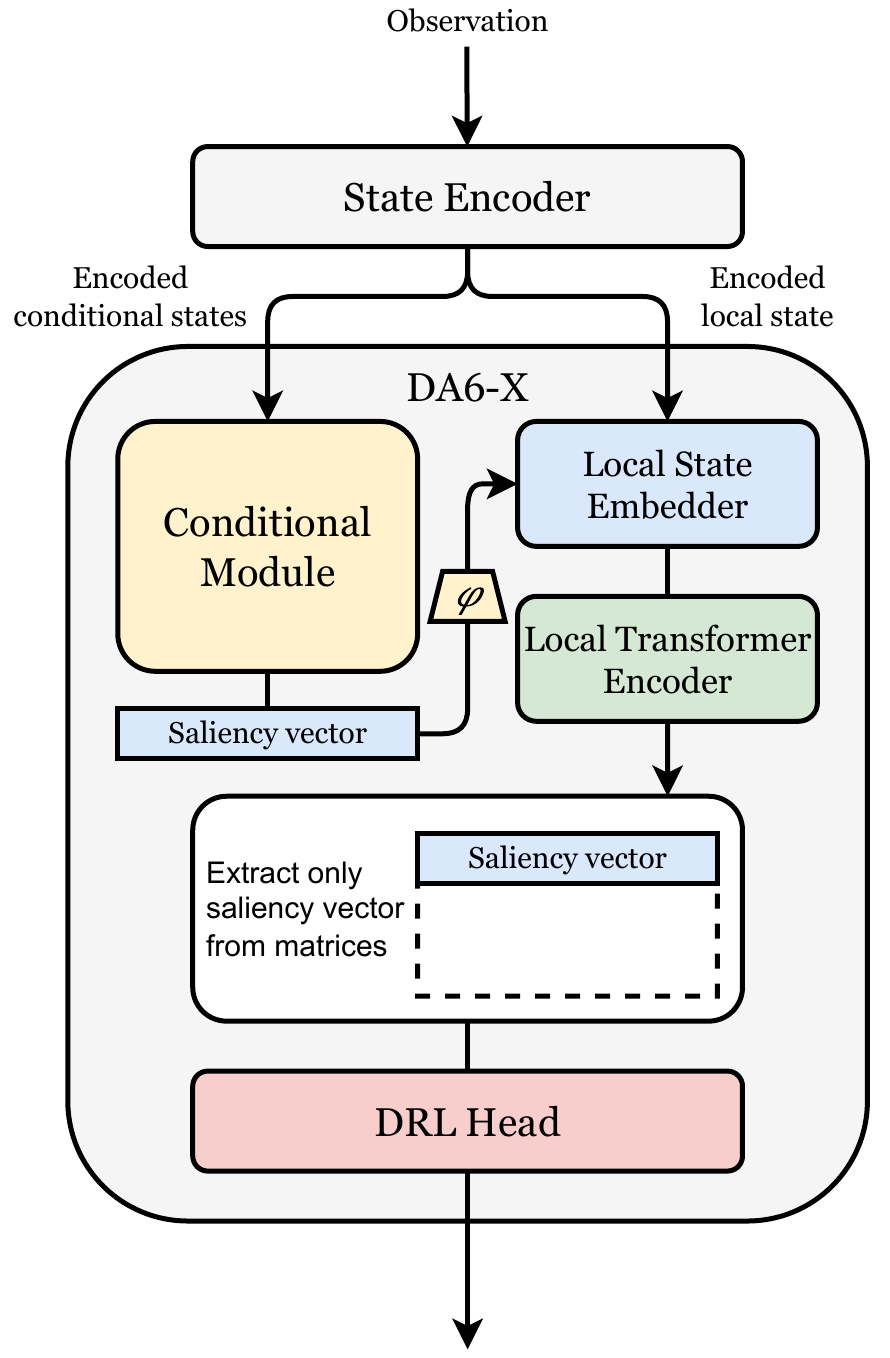}\\
    \subcaption{DA6-X.}\label{fig:DA6}
  \end{minipage}
  \caption{Architecture of baseline and proposed method.}\label{fig:proposal}
\end{figure*}

The key idea of DA6-X is to consistently use the same saliency vector in CM and the local transformer encoder, as illustrated in Fig.~\ref{fig:proposal}. CM and the local transformer encoder play the roles of recognizing the environmental conditions and weighing the local information by the attention mechanism, respectively. DA6-X is an extension of DA3-X; however, it is different in that DA6-X incorporates CM to explain the rationale for situation-dependent behaviors and to facilitate the understanding of administrators. We incorporated separate transformer encoders in CM to effectively generate the saliency vector. Through the attention mechanism in the transformer encoders in CM, DA6-X can flexibly recognize the conditional states and represent them in the saliency vector that is eventually reused in the local transformer encoders. We can clarify what CM does because it is also interpretable by analyzing the attention weights in the transformer encoders in CM. Similar to the attention analysis demonstrated in Section~\ref{sec:attention}, further attention analysis of how DA6-X handles the conditional states in CM can be similarly performed. The mathematical procedure of DA6-X is described as follows:
\par
\begin{equation}
  \begin{split}\nonumber
    g_{m,0} &= [\sal{u}_m; x^1_m E^\mathrm{cond}_m; x^2_m E^\mathrm{cond}_m; \ldots ; x^{I_m}_m E^\mathrm{cond}_m] + P^{\mathrm{cond}}_m\\
    g_{m,l} &= \mathrm{TEL}^\mathrm{cond}_m (g_{m, l-1}), \qquad l = 1, \ldots , L_m\\
    \sal{v} &= \mathrm{VectorIntegration}(g^0_{1,L_1}, \ldots, g^0_{m,L_m}, \ldots, g^0_{M,L_M})
  \end{split}
\end{equation}
The equation above expresses the flow calculations of the conditional states in CM. Suppose there are $M$ submodules in total; the $m$-th $(1 \leq m \leq M)$ patched conditional state matrix $x_m \in \mathbb{R}^{I_m \times (P^2_m N_m)}$ is embedded by parameters $E^\mathrm{cond}_m \in \mathbb{R}^{(P^2_m N_m) \times C_m}$ and subsequently concatenated with the saliency vector (trainable parameters) $\sal{u}_m \in \mathbb{R}^{C_m}$, where $I_m$ is the length of the $m$-th conditional state after the patched operation; $P_m$ is the $m$-th patch size; $N_m$ is the number of conditional state channels before the patched operation; and $C_m$ is the length of the $m$-th saliency vector. Note that this calculation corresponds to the procedure in the \emph{state embedder}, as shown in Fig.~\ref{fig:ConditionalModule}. After positional embedding $P^{\mathrm{cond}} \in \mathbb{R}^{(I_m +1) \times C_m}$ is appended, the $m$-th input is fed into the $m$-th transformer encoder layer $(\mathrm{TEL}^\mathrm{cond}_m)$, which is illustrated by the green box in Fig.~\ref{fig:ConditionalModule}. After looping $L_m$ times, saliency vector $g^0_{1,L_1}, \ldots, g^0_{m,L_m}, \ldots, g^0_{M,L_M}$ from all submodules are aggregated in a \emph{vector integration} procedure to produce the final saliency vector $\sal{v} \in \mathbb{R}^{C_\mathrm{L}}$ of length $C_\mathrm{L}$.
\par

The calculation after CM is described as follows:
\begin{equation}
  \begin{split}\nonumber
    h_0 &= [\varphi(\sal{v}); y^1 E^\mathrm{local}; y^2 E^\mathrm{local}; \ldots ; y^{I_\mathrm{L}} E^\mathrm{local}] + P^{\mathrm{local}}\\
    h_l &= \mathrm{TEL}^{\mathrm{local}}(h_{l-1}), \qquad l = 1, \ldots , L_\mathrm{L}\\
    Q &= \mathrm{DRLHead}(h^0_{L_\mathrm{L}})
  \end{split}
\end{equation}
After the projection denoted by $\varphi(\cdot)$, the saliency vector $\sal{v}$ is directly fed into the \emph{local state embedder}, as shown in Fig.~\ref{fig:DA6}. Similar to the previous process, the patched local observation $y \in \mathbb{R}^{I_\mathrm{L} \times (P^2_\mathrm{L} N_\mathrm{L})}$ is embedded by parameters $E_\mathrm{L} \in \mathbb{R}^{(P^2_\mathrm{L} N_\mathrm{L}) \times C_\mathrm{L}}$ and concatenated to the projected saliency vector $\varphi(\sal{v})$, where $I_\mathrm{L}$ is the length of local observation after the patched operation; $P_\mathrm{L}$ is the local patch size; and $N_\mathrm{L}$ is the number of patched local observation channels. Reusing the saliency vector $\sal{v}$ of CM, DA6-X agents can build flexible policies after considering the conditional states. Similarly, the input is forwarded to the local transformer encoder layer $(\mathrm{TEL}^{\mathrm{local}})$ $L_\mathrm{L}$ times after the positional embedding $P^{\mathrm{local}} \in \mathbb{R}^{(I_\mathrm{L}+1)\times C_\mathrm{L}}$ is appended. Finally, only the saliency vector $h^0_{L_\mathrm{L}}$ is fed into \emph{DRLHead}, as shown in Fig.~\ref{fig:DA6}.
\par

\subsection{Advantages of DA6-X}
Because of the selection ability of the attention mechanism, agents achieve higher efficiency in exploration and even better robust coordination than when the attention mechanism is not incorporated in the system. Hence, DA6-X agents with the attention mechanism incorporated are also expected to outperform the conventional MADRL or DRL algorithms, similar to the previous method~\cite{MAT_DQN,DA3}. In addition, the output of DRLHead denoted as $Q$ in the equation of DA6-X can be Q-values or a stochastic policy, depending on the DRL algorithm to incorporate, such as DDPG~\cite{Lillicrap2016ContinuousCW}, PPO~\cite{schulman2017ppo}, or RAINBOW~\cite{RAINBOW}. This feature also leads to higher versatility of DA6-X in real-world applications.
\par

The reuse of the saliency vector in DA6-X is beneficial not only for performance improvement but also for improved interpretability when compared to DA3-X~\cite{DA3}. As explained earlier, DA6-X agents can flexibly change their strategy of weighing a particular piece of information in their local observation depending on an arbitrary number of available conditional states by reusing the saliency vector. Hence, complex behaviors of DA6-X agents can be interpretable using attention heatmaps in their local observation. By contrast, DA3-X agents directly feed the aggregated conditional states to the black-box DRLHead, such that their interpretability in local observations is relatively limited.
\par

The main difference between our proposed method, DA6-X, and DA3-X is CM. Indeed, DA3-X can be regarded as DA6-X without CM ($M = 0$) as shown in Fig.~\ref{fig:DA3}. Reusing the saliency vector from CM in the latter local transformer encoder, it is interpretable how agents see their local observation after they concern their conditional states in CM, such as their global positions (\emph{G pos}). Therefore, because of the lack of its CM, DA3-X always acts in the same manner and its interpretability is consistent, whenever its local observation is the same and the conditional states vary. In contrast, DA6-X can flexibly change its way of obtaining the local observation after understanding the conditional states, such that it may act differently depending on the conditional states even though the same local observation is obtained. Because the saliency vector is reused, DA6-X possesses improved interpretability than that of DA3-X.
\par

\section{Experiments and Results}
\subsection{Experimental Setup}\label{sec:ExperimentalSetup}
Our experiment was conducted using the objects' collection game in an environment $\env$, as shown in Fig.~\ref{fig:environment}, where $(G_X, G_Y) = (25, 25)$, and each color of the cell indicates the entity: black for walls, blue for initial position of agents, white for empty, and green, red, and beige for object spawn regions. Two types of objects were spawned in $\env$: objects $\bigstar$ in green regions, objects of type $\blacktriangle$ in red regions, and both objects in the beige region. The number of objects was set to $20$ for each type ($40$ objects in total). The environment was divided into four distinct sections ($\Gamma$, $\Delta$, $\Theta$, and $\Lambda$), as shown in Fig.~\ref{fig:map_environment2}. Once an agent collected an object, the same type of object was spawned at random location within the spawn region.
\par
\begin{wraptable}[9]{r}{50mm}
  \vspace{-0.2\baselineskip}
  %  \centering
   \caption{Agent-task specification.}\label{table:agent-spec}
   \begin{tabular*}{50mm}{lccl}
     \toprule
     \small{Agents} & \small{Num} & \small{Symbol} & \small{Region}\\
     \midrule
     \small{Type A} & 2 & $\bigstar$ & $\Gamma$, $\Lambda$\\
     \small{Type B} & 2 & $\blacktriangle$ & $\Delta$, $\Theta$\\
     \small{Type C} & 2 & $\bigstar$, $\blacktriangle$ & $\Gamma$, $\Delta$\\
     \small{Type D} & 2 & $\bigstar$, $\blacktriangle$ & $\Theta$, $\Lambda$\\
     \hline
   \end{tabular*}
 \end{wraptable}
Eight agents ($N=8$) were placed at blue cells and learned their individual policies $\pi_i$ to collect as many objects as possible without colliding for $5,000$ episodes, where the episode length was $H=200$ steps. These agents were classified into four types, as shown in Table~\ref{table:agent-spec}, which lists the agent type, number of agents, assigned object type, and region to collect. For example, there are two \emph{type A} agents which collect $\bigstar$ objects in the diagonal regions ($\Gamma$ and $\Lambda$ in Fig.~\ref{fig:map_environment2}), and two \emph{type C} agents which collect both $\bigstar$ and $\blacktriangle$ objects only in the top-half region ($\Gamma$ and $\Delta$ in Fig.~\ref{fig:map_environment2}). In other words, cooperative agents can be non-cooperative in another region. The reward scheme was set as $r_\mathrm{o}=1$ and $r_\mathrm{c}=-1$, where $r_\mathrm{o}$ is attributed to an agent only when it collects the assigned type object in the allocated region in Table~\ref{table:agent-spec}.
\par

The purpose of our experiment was to analyze the attention weights in the local transformer encoders to investigate how DA6-X agents build situation-dependent coordination when encountering various conditional states and how these conditional states affect their behavioral decisions. We examined eight \emph{DA6-DQN} agents and \emph{DA6-IQN} agents, which had the MLP and \emph{implicit quantile network} (IQN)~\cite{IQN} installed in their DRLHead, respectively. For baseline methods, eight standard DQN agents, IQN agents, DA3-DQN agents, and DA3-IQN agents were trained. In addition to the local observation that was fed into the local state encoder, two types of additional input were provided to the CM: the global position of the observing agent (\emph{G pos}) and all objects' positions (\emph{O pos}) for the conditional states in the environment. Each agent obtains its local observation ($\{R_\mathrm{X}, R_\mathrm{Y}\} = \{7, 7\}$) from \emph{local view}~\cite{LocalView}. The baseline agents observe \emph{G pos} and \emph{O pos} by \emph{relative view}~\cite{LocalView}, while the DA6-X agents obtain the conditional states by \emph{merged view}~\cite{MergedView}. The experiment was repeated three times for each condition by using approximately $100$ GPU hours of \emph{NVIDIA GeForce RTX 3090}.
\par

\begin{figure*}
  \centering
  \begin{minipage}[t]{0.48\hsize}
    \centering
    \includegraphics[keepaspectratio, width=\linewidth]{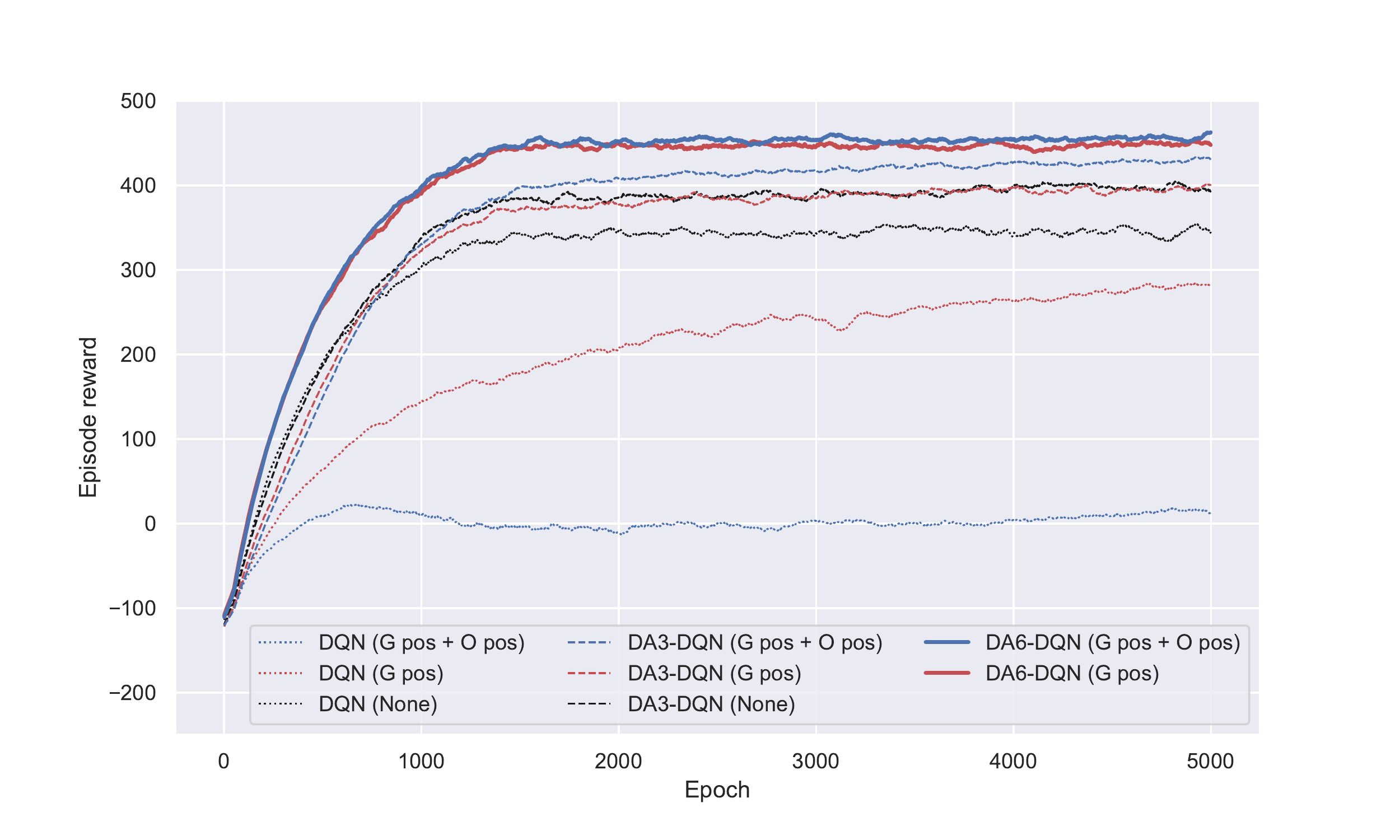}\\
    \subcaption{Episode reward by DQN-based algorithms.}
  \end{minipage}
  \hfil
  \begin{minipage}[t]{0.48\hsize}
    \centering
    \includegraphics[keepaspectratio, width=\linewidth]{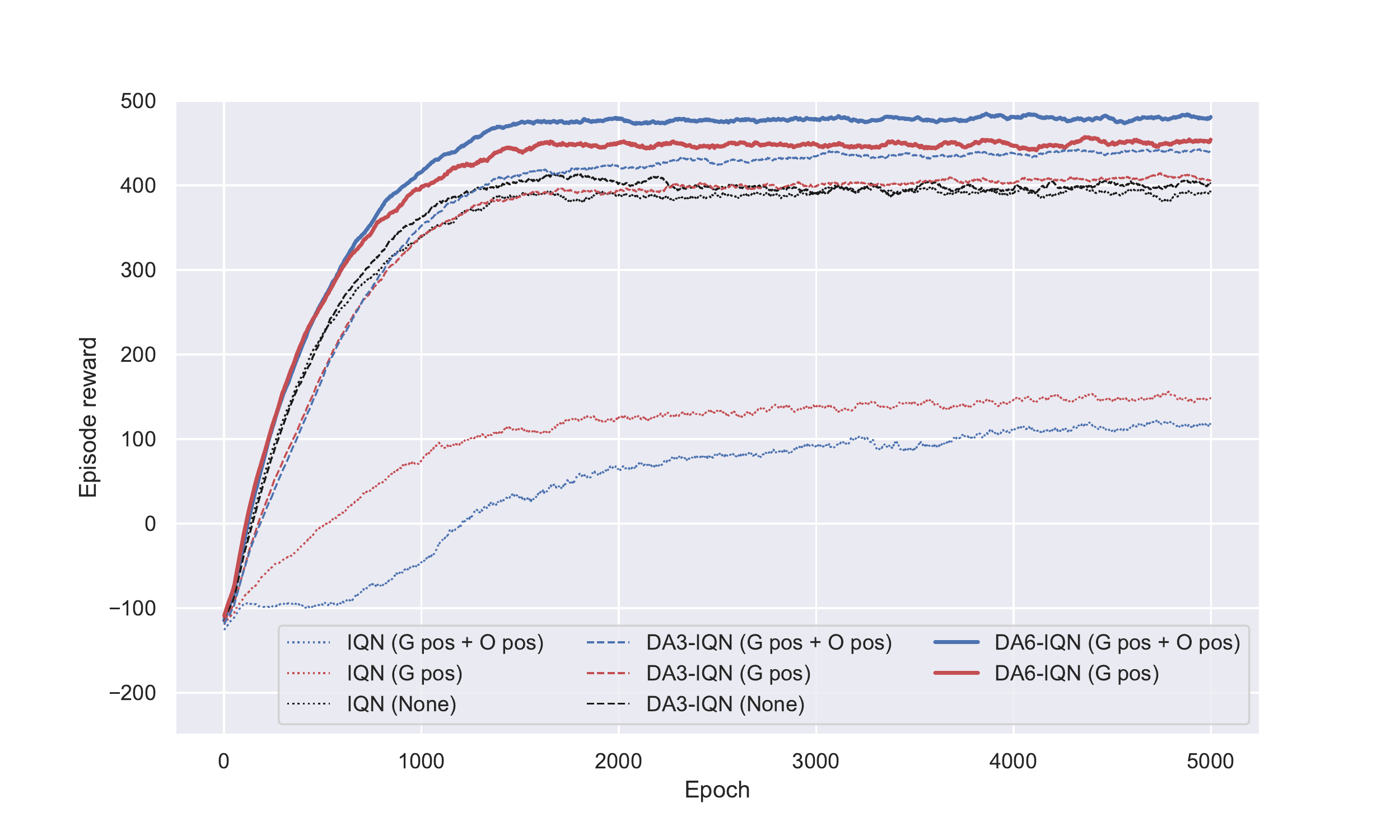}\\
    \subcaption{Episode reward by IQN-based algorithms.}
  \end{minipage}
  \caption{Learning performance comparison with varying conditional states.}\label{fig:performance-result}
\end{figure*}

\begin{table*}[t]
  \centering
  \caption{Quantitative performance comparison.}\label{table:quantitative-result}
  \small
  \begin{tabular}{llllll}
    \toprule
    Input to CM & Model & Episode reward & Objects counts & Agents collision & Walls collision\\
    \midrule
    None & DQN & $348.64 \pm 35.83$ & $360.22 \pm 35.02$ & $6.35 \pm 2.50$ & $5.23 \pm 2.73$\\
    & DA3-DQN & $394.76 \pm 34.92$ & $402.43 \pm 35.07$ & $3.94 \pm 1.40$ & $3.72 \pm 1.29$\\
    & IQN & $389.94 \pm 28.43$ & $398.85 \pm 28.48$ & $4.30 \pm 1.57$ & $4.61 \pm 1.71$\\
    & DA3-IQN & $402.16 \pm 30.44$ & $410.41 \pm 30.43$ & $4.10 \pm 1.33$ & $4.15 \pm 1.69$\\
    \midrule
    \emph{G pos} & DQN & $282.21 \pm 22.75$ & $305.59 \pm 21.26$ & $14.24 \pm 6.71$ & $9.14 \pm 4.32$\\
    & DA3-DQN & $397.94 \pm 25.43$ & $410.95 \pm 25.82$ & $7.62 \pm 3.68$ & $5.38 \pm 3.32$\\
    & DA6-DQN & $\mathbf{449.69 \pm 24.96}$ & $\mathbf{457.08 \pm 25.13}$ & $\mathbf{3.84 \pm 1.21}$ & $\mathbf{3.55 \pm 1.37}$\\
    & IQN & $147.77 \pm 27.12$ & $323.51 \pm 19.75$ & $13.78 \pm 7.45$ & $161.97 \pm 13.42$\\
    & DA3-IQN & $407.29 \pm 25.69$ & $419.39 \pm 24.18$ & $6.55 \pm 2.85$ & $5.55 \pm 5.23$\\
    & DA6-IQN & $\mathbf{452.64 \pm 25.91}$ & $\mathbf{460.65 \pm 25.73}$ & $\mathbf{4.09 \pm 1.40}$ & $\mathbf{3.92 \pm 1.68}$\\
    \midrule
    \emph{G pos} + \emph{O pos} & DQN & $14.60 \pm 21.32$ & $73.50 \pm 10.90$ & $39.36 \pm 11.93$ & $19.54 \pm 6.54$\\
    & DA3-DQN & $432.20 \pm 22.96$ & $446.11 \pm 22.33$ & $ 8.46 \pm 3.36$ & $5.45 \pm 2.72$\\
    & DA6-DQN & $\mathbf{456.05 \pm 24.28}$ & $\mathbf{463.81 \pm 24.37}$ & $\mathbf{4.04 \pm 1.39}$ & $\mathbf{3.72 \pm 1.10}$\\
    & IQN & $117.73 \pm 29.48$ & $342.13 \pm 15.20$ & $19.01 \pm 10.28$ & $205.40 \pm 19.35$\\
    & DA3-IQN & $441.15 \pm 19.98$ & $455.87 \pm 18.72$ & $8.37 \pm 3.73$ & $6.34 \pm 4.40$\\
    & DA6-IQN & $\mathbf{479.61 \pm 20.30}$ & $\mathbf{488.96 \pm 19.76}$ & $\mathbf{5.30 \pm 1.48}$ & $\mathbf{4.05 \pm 1.47}$\\
    \hline
  \end{tabular}
\end{table*}

\subsection{Performance Comparison}\label{sec:performance}
Figure~\ref{fig:performance-result} shows the averaged episode reward over $5,000$ episodes for each reinforcement learning method. The color of lines in Fig.~\ref{fig:performance-result} are indicatve of the conditional states fed to agents: blue for both \emph{G pos} and \emph{O pos}, red for \emph{G pos}, and black for no conditional states. Table~\ref{table:quantitative-result} lists the averaged episode reward, collected number of objects, and collisions that occurred between agents and walls in the final $100$ episode for each learning algorithm and conditional states to CM. In general, learning performance improved when more input information was available during decision-making regardless of the DRL method, and similar phenomena were confirmed except the DQN and IQN agents.
\par

{\bfseries When \emph{G pos} is available:}
According to Fig.~\ref{fig:performance-result} and Table~\ref{table:quantitative-result}, providing the location of observing agents (\emph{G pos}) resulted in DA3-X and DA6-X agents collecting more objects. DA3-DQN and DA3-IQN agents with \emph{G pos} collect $8.52$ ($2.12\%$) and $8.98$ ($2.19\%$) more objects when compared with DA3-DQN and DA3-IQN agents, which had no input (None) to CM, while DA6-DQN and DA6-IQN agents with \emph{G pos} obtain $54.65$ ($13.58\%$) and $50.24$ ($12.24\%$) more objects, thereby achieving $54.93$ and $50.48$ points more in episode rewards. In addition, DA6-DQN and DA6-IQN agents occurred only $7.39$ and $8.01$ collisions in total, which are $43.15\%$ and $33.80\%$ less than those of DA3-DQN and DA3-IQN agents with \emph{G pos}. Performance degradation is confirmed for the standard DQN and IQN agents with \emph{G pos} because of its state complexity overload.
\par

{\bfseries When \emph{G pos} and \emph{O pos} are available:}
We confirmed that DA3-X and DA6-X agents achieved even higher learning performance when the location of both observing agents and objects (\emph{G pos} + \emph{O pos}) were available. As shown in Table~\ref{table:quantitative-result}, DA3-DQN and DA3-IQN agents with \emph{G pos} and \emph{O pos} achieve $37.44$ and $38.99$ points more in episode rewards when compared with those of DA3-DQN and DA3-IQN agents without conditional states, while DA6-DQN and DA6-IQN agents improve $61.29$ and $77.45$ points in episode reward. Moreover, the differences in collected objects by DA3-DQN versus DA6-DQN agents and DA3-IQN versus DA6-IQN are $17.69$ ($3.97\%$) and $33.09$ ($7.26\%$), confirming that DA6-X agents successfully build their efficient policy with the conditional states. Similar to the previous case, learning performance by the standard DQN and IQN agents gets worse with \emph{G pos} and \emph{O pos}.
\par

\subsection{Attention and Coordination}\label{sec:attention}
It was confirmed that providing conditional states makes DA6-X agents learn more efficient policies. The manner in which the conditional states affect the decision-making process was further investigated to improve the overall performance as well as for successful coordination by analyzing the intensity of the attention weights in particular situations. The coordination analysis was conducted for two cases: when only \emph{G pos} was available and when \emph{G pos} and \emph{O pos} were available.
\par

\begin{figure}[t]
  \begin{minipage}[t]{0.48\hsize}
    \centering
    \begin{minipage}[t]{0.48\hsize}
      \centering
      \includegraphics[keepaspectratio, width=\linewidth]{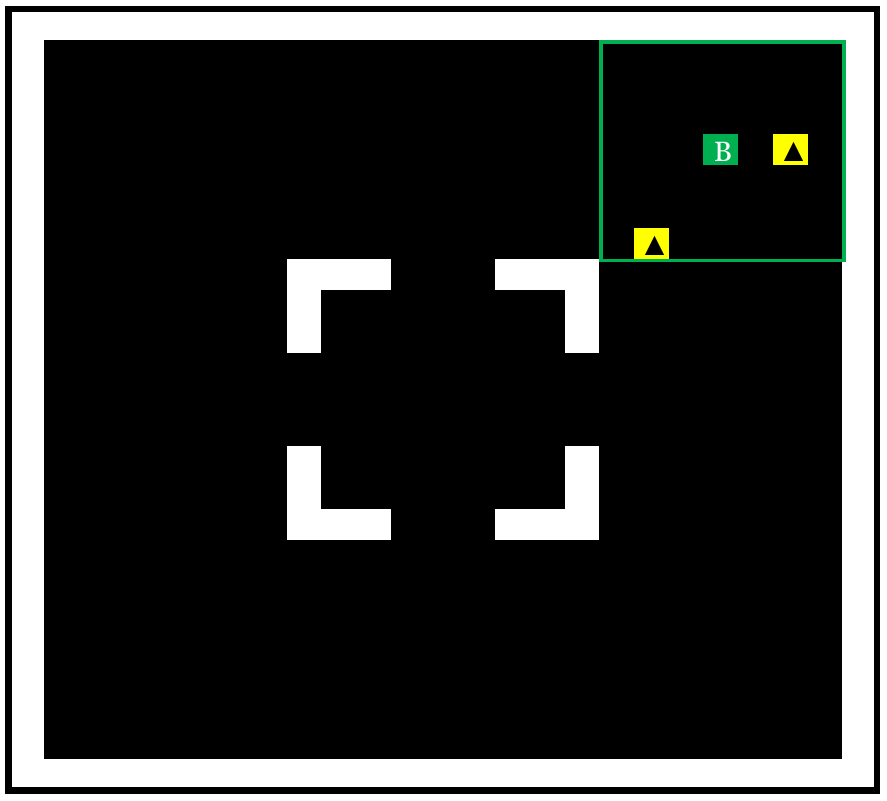}
    \end{minipage}
    \begin{minipage}[t]{0.48\hsize}
      \centering
      \includegraphics[keepaspectratio, width=\linewidth]{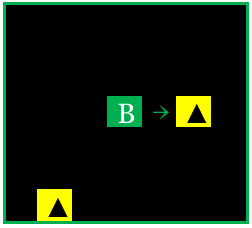}
    \end{minipage}\\
    \begin{minipage}[t]{0.48\hsize}
      \centering
      \includegraphics[keepaspectratio, width=\linewidth]{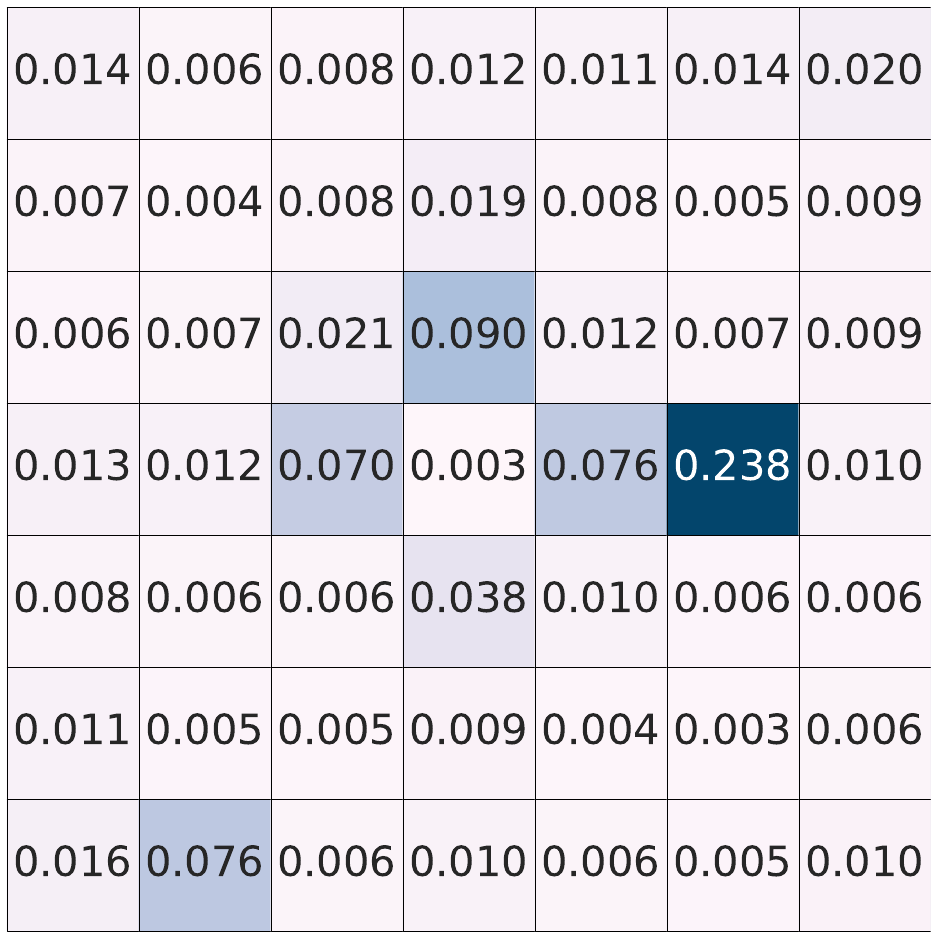}
    \end{minipage}
    \begin{minipage}[t]{0.48\hsize}
      \centering
      \includegraphics[keepaspectratio, width=\linewidth]{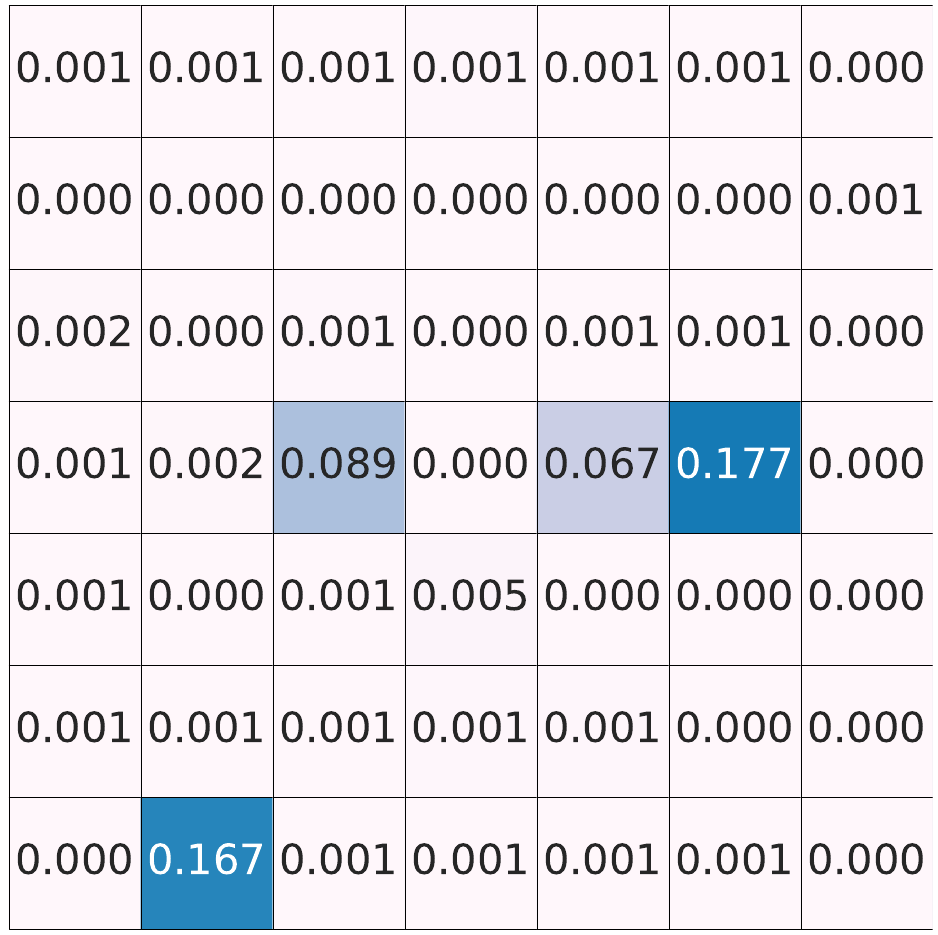}
    \end{minipage}\\
    \subcaption{Solo in $\Delta$.}\label{fig:analysis-Gpos-A}
  \end{minipage}
  \begin{minipage}[t]{0.48\hsize}
    \centering
    \begin{minipage}[t]{0.48\hsize}
      \centering
      \includegraphics[keepaspectratio, width=\linewidth]{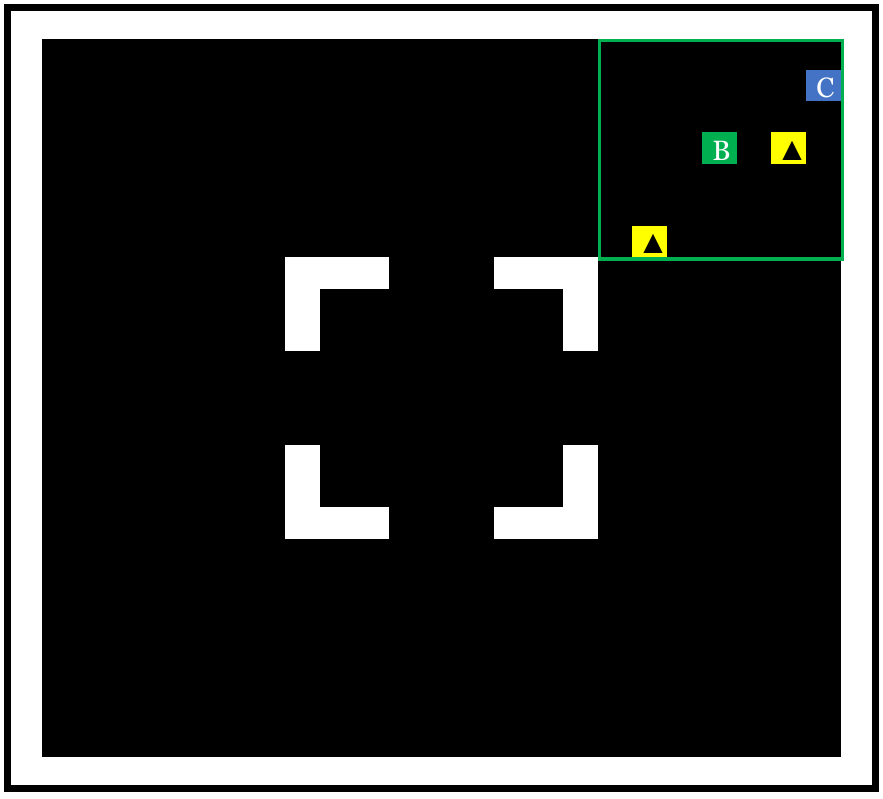}
    \end{minipage}
    \begin{minipage}[t]{0.48\hsize}
      \centering
      \includegraphics[keepaspectratio, width=\linewidth]{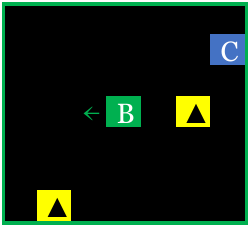}
    \end{minipage}\\
    \begin{minipage}[t]{0.48\hsize}
      \centering
      \includegraphics[keepaspectratio, width=\linewidth]{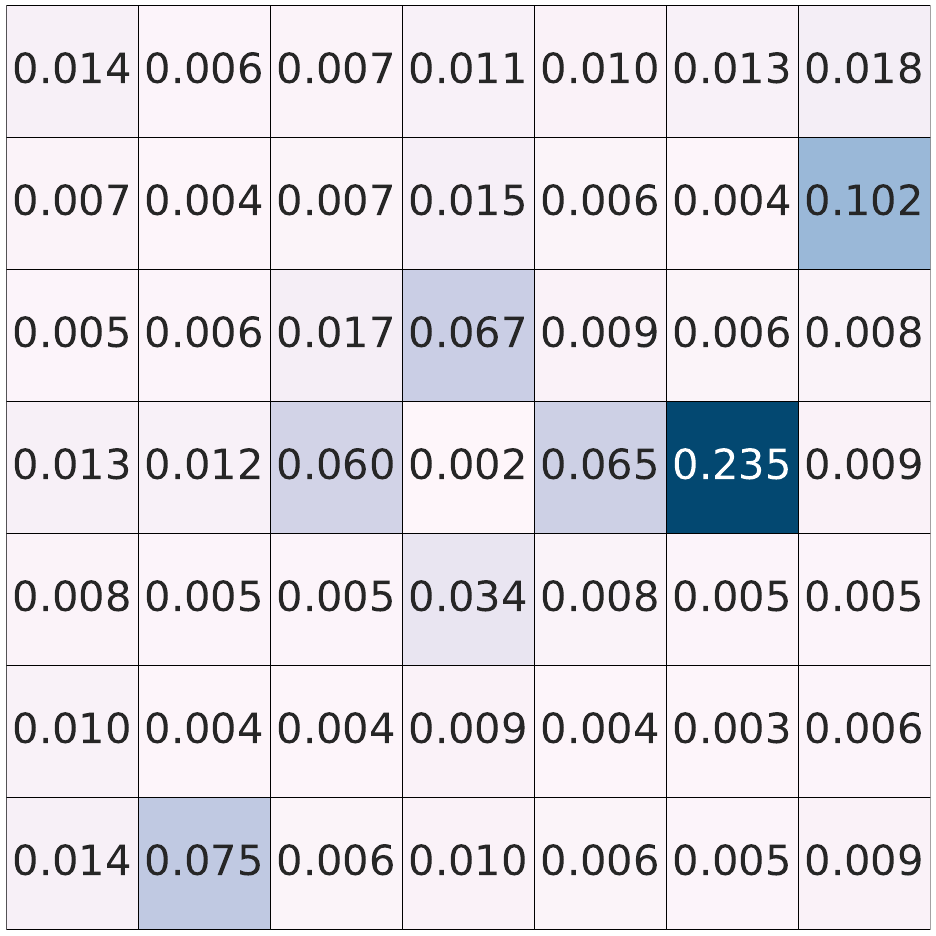}
    \end{minipage}
    \begin{minipage}[t]{0.48\hsize}
      \centering
      \includegraphics[keepaspectratio, width=\linewidth]{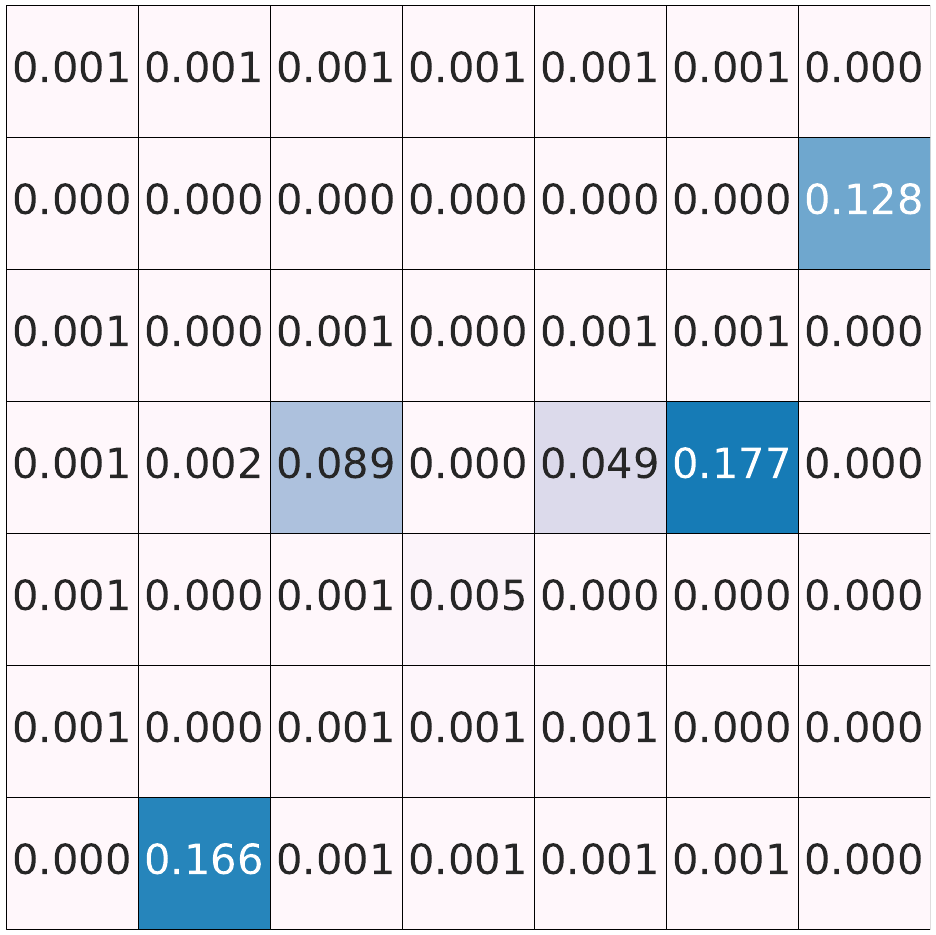}
    \end{minipage}\\
    \subcaption{With \emph{type C} agent in $\Delta$.}\label{fig:analysis-Gpos-B}
  \end{minipage}
  \vspace{0.5\baselineskip}\\
  \begin{minipage}[t]{0.48\hsize}
    \centering
    \begin{minipage}[t]{0.48\hsize}
      \centering
      \includegraphics[keepaspectratio, width=\linewidth]{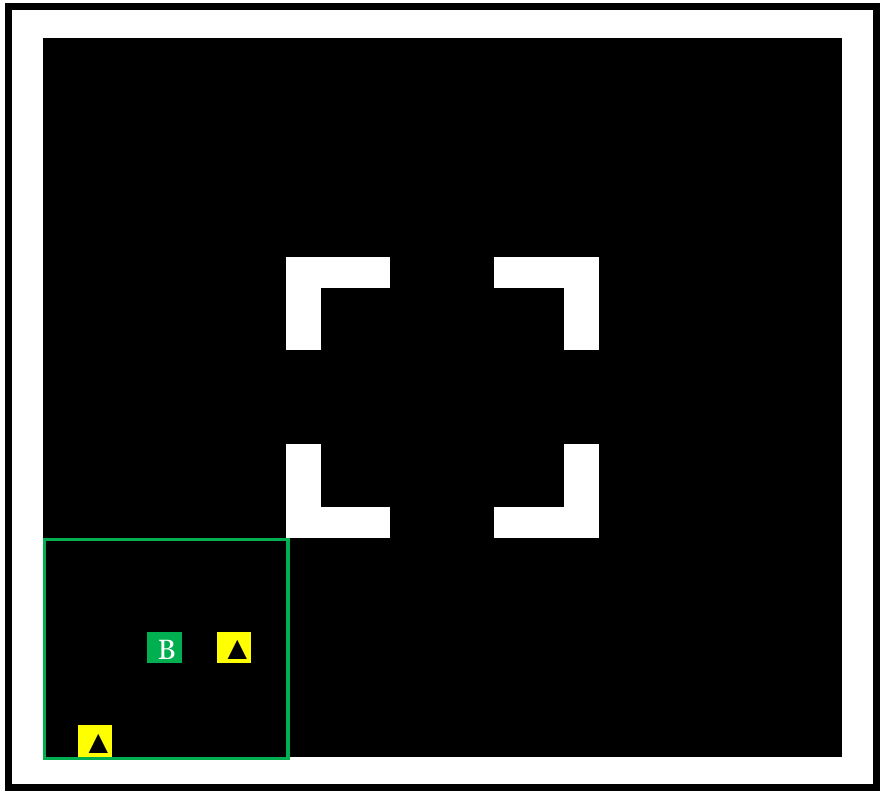}
    \end{minipage}
    \begin{minipage}[t]{0.48\hsize}
      \centering
      \includegraphics[keepaspectratio, width=\linewidth]{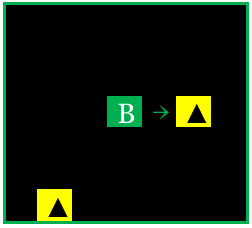}
    \end{minipage}\\
    \begin{minipage}[t]{0.48\hsize}
      \centering
      \includegraphics[keepaspectratio, width=\linewidth]{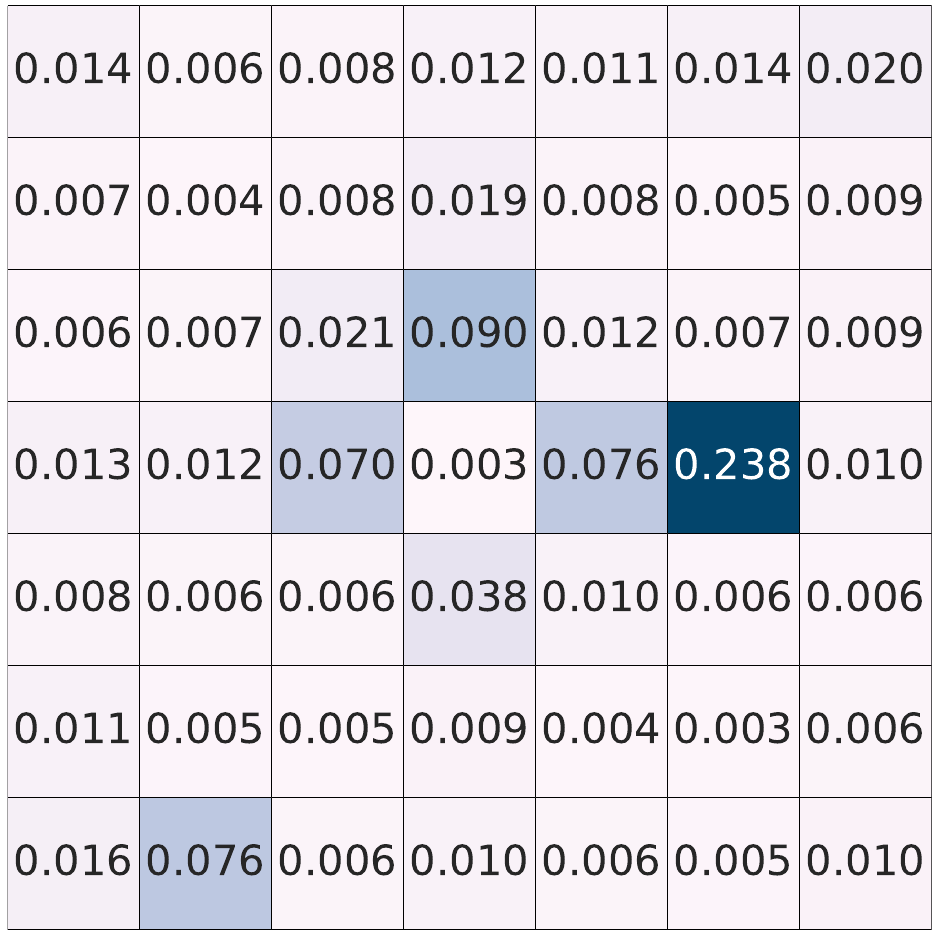}
    \end{minipage}
    \begin{minipage}[t]{0.48\hsize}
      \centering
      \includegraphics[keepaspectratio, width=\linewidth]{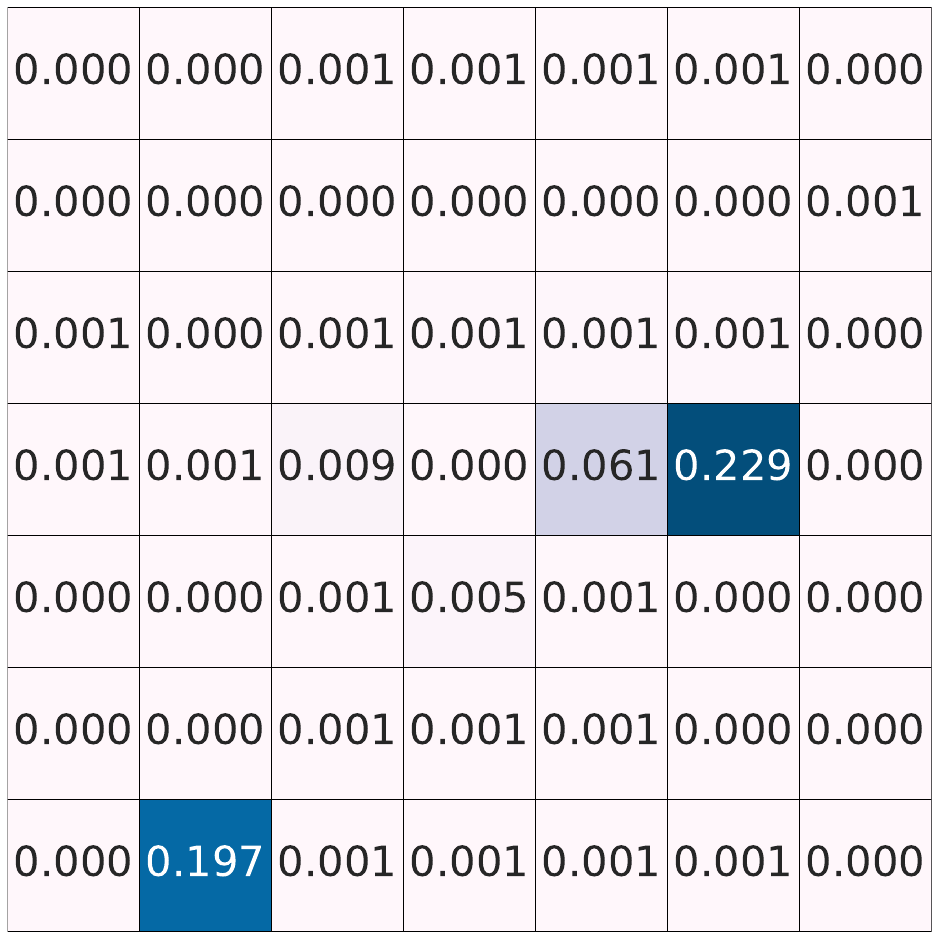}
    \end{minipage}\\
    \subcaption{Solo in $\Theta$.}\label{fig:analysis-Gpos-C}
  \end{minipage}
  \begin{minipage}[t]{0.48\hsize}
    \centering
    \begin{minipage}[t]{0.48\hsize}
      \centering
      \includegraphics[keepaspectratio, width=\linewidth]{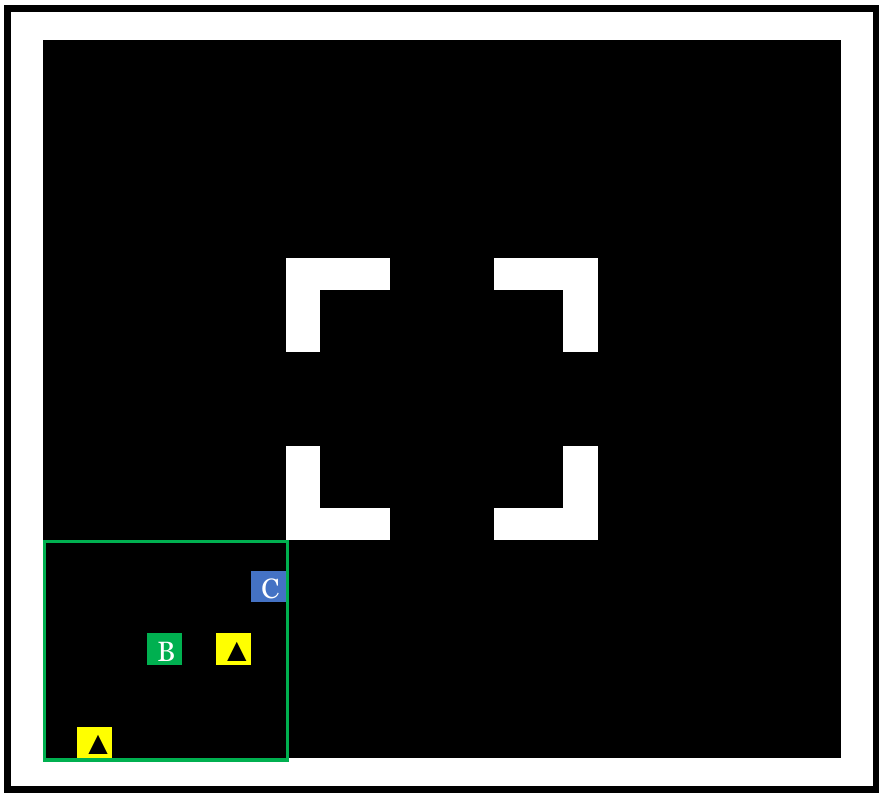}
    \end{minipage}
    \begin{minipage}[t]{0.48\hsize}
      \centering
      \includegraphics[keepaspectratio, width=\linewidth]{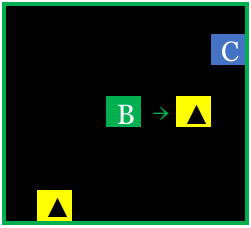}
    \end{minipage}\\
    \begin{minipage}[t]{0.48\hsize}
      \centering
      \includegraphics[keepaspectratio, width=\linewidth]{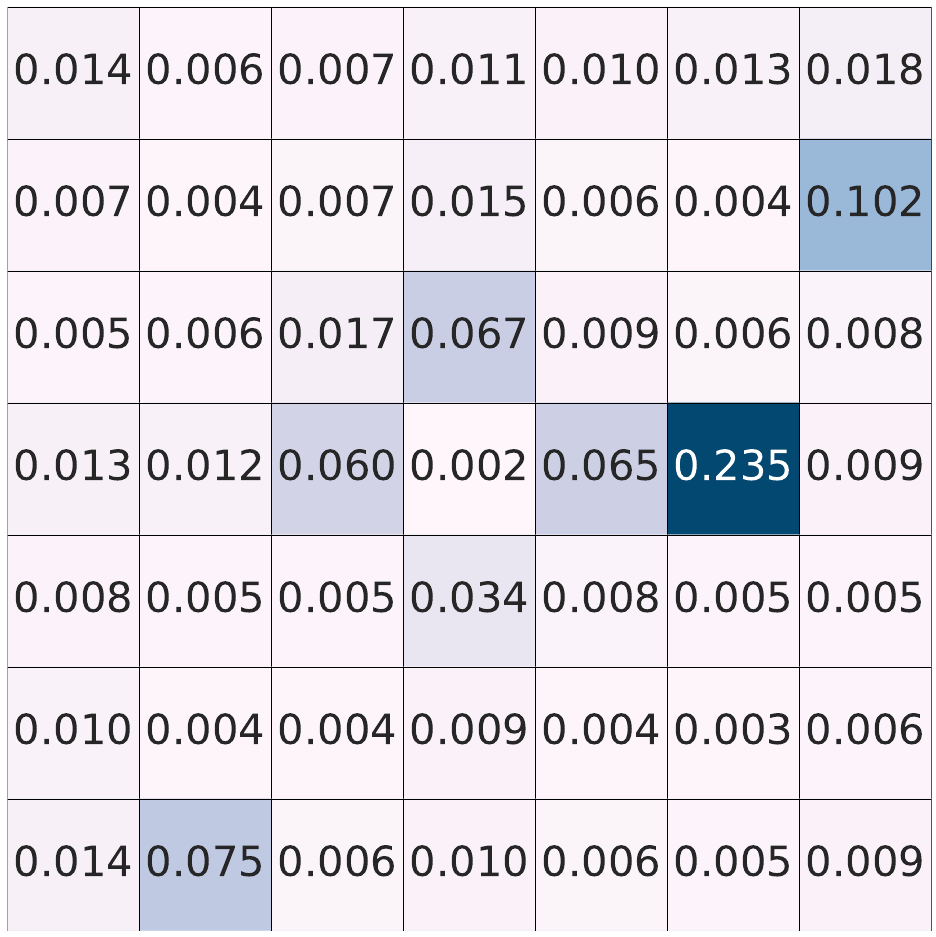}
    \end{minipage}
    \begin{minipage}[t]{0.48\hsize}
      \centering
      \includegraphics[keepaspectratio, width=\linewidth]{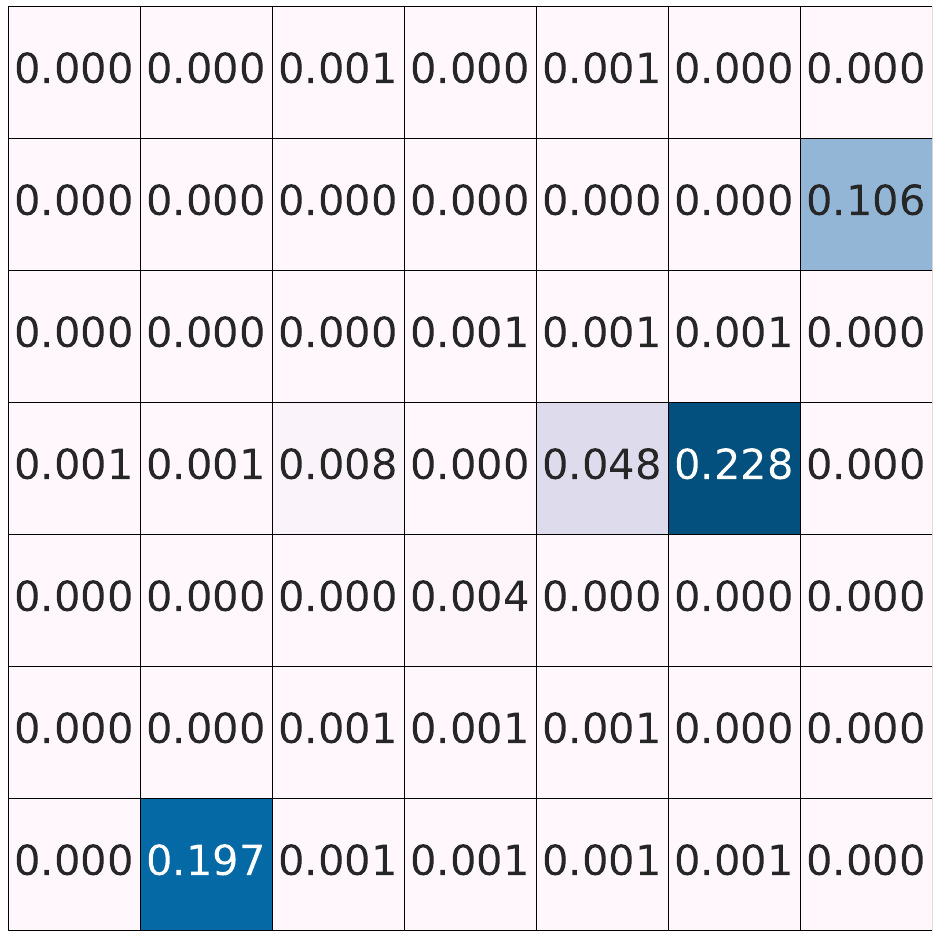}
    \end{minipage}\\
    \subcaption{With \emph{type C} agent in $\Theta$.}\label{fig:analysis-Gpos-D}
  \end{minipage}
  \caption{Coordination study with \emph{G pos}.}\label{fig:analysis-Gpos}
\end{figure}

{\bfseries When \emph{G pos} is available:}
DA6-DQN agents improved their performance by understanding their global position via the saliency vector. Hence, the way the coordinates vary was analyzed by testing at two different locations: right-up and left-bottom regions ($\Delta$ and $\Theta$). First, how a \emph{type B} agent behaved with two objects at different distances was observed with/without \emph{type C} agent located nearby, as depicted in Fig.~\ref{fig:analysis-Gpos}. Each figure shows the global position of the \emph{type B} agent (top left), the local observation (top right), and the corresponding attention heatmaps of DA3-DQN (bottom left) and DA6-DQN (bottom right) for interpretability comparison. Notably, the attention heatmap of DA3-DQN is generated from the attention weights of the transformer encoder in Fig.~\ref{fig:DA3} without conditional states, while that of DA6-DQN from the local transformer encoder in Fig.~\ref{fig:DA6}. The green node in the global position (Fig.~\ref{fig:analysis-Gpos}) is the observing agent itself, and the green square expresses the visible region ($\{R_\mathrm{X}, R_\mathrm{Y}\} = \{7, 7\}$). Color in the observation indicates the entities in the environment: white for walls, black for empty, green for observing the \emph{type B} agent, yellow for the $\blacktriangle$ object, blue for the \emph{type C} agent. Green arrows in the local observation indicate the orientation of next movement decided by DA6-DQN.
\par

According to Fig.~\ref{fig:analysis-Gpos-A} and \ref{fig:analysis-Gpos-C}, the \emph{type B} agent always approaches the closer object by moving toward the right as the closer object attracts more attention than the one farther away ($0.177$ and $0.089$ of DA6-DQN attention when the agent is in the right-upper region $\Delta$ whereas $0.229$ and $0.197$ of DA6-DQN attention when it is in the left-bottom region $\Theta$) regardless of the global position. The attention heatmap by DA3-DQN also puts higher attention on a closer object ($0.238$) than a farther object ($0.076$). However, the attention values are exactly the same in both Fig.~\ref{fig:analysis-Gpos-A} and \ref{fig:analysis-Gpos-C} because DA3-X does not consider conditional states.
\par

Interestingly, when the \emph{type C} agent is located around the closer object (Fig.~\ref{fig:analysis-Gpos-B} and \ref{fig:analysis-Gpos-D}), the \emph{type B} agent behaves differently depending on its global position. In the right-upper region $\Delta$ (Fig.~\ref{fig:analysis-Gpos-B}), the \emph{type B} agent moves leftward to approach the object farther out by yielding to the \emph{type C} agent as the attention weight of DA6-DQN is $0.128$. The \emph{type B} agent understands that the \emph{type C} agent is aiming at the same target, interpreting it as worth building coordination to avoid competition. In contrast, at the left-bottom region $\Theta$ (Fig.~\ref{fig:analysis-Gpos-D}), the \emph{type B} agent assigns an attention weight of DA6-DQN at only $0.106$ to the \emph{type C} agent and approaches the closer object by moving rightward. The \emph{type C} agent cannot collect the object in $\Theta$, which is outside its assigned task region according to Table~\ref{table:agent-spec}. The \emph{type B} agent successfully learned that it is not worth building coordination with the \emph{type C} agent at the left-bottom region $\Theta$ through training experience. This policy of \emph{type B} agent is verified from the reduction in attention weight (from $0.128$ to $0.106$ of DA6-DQN attention) on the \emph{type C} agent at two different locations. Moreover, the behavioral analysis of how the conditional states influence the cooperative behaviors by DA3-X agents becomes very complicated because the attention heatmap by DA3-DQN does not change depending on the global position.
\par

\begin{figure}[t]
  \begin{minipage}[t]{0.48\hsize}
    \centering
    \begin{minipage}[t]{0.48\hsize}
      \centering
      \includegraphics[keepaspectratio, width=\linewidth]{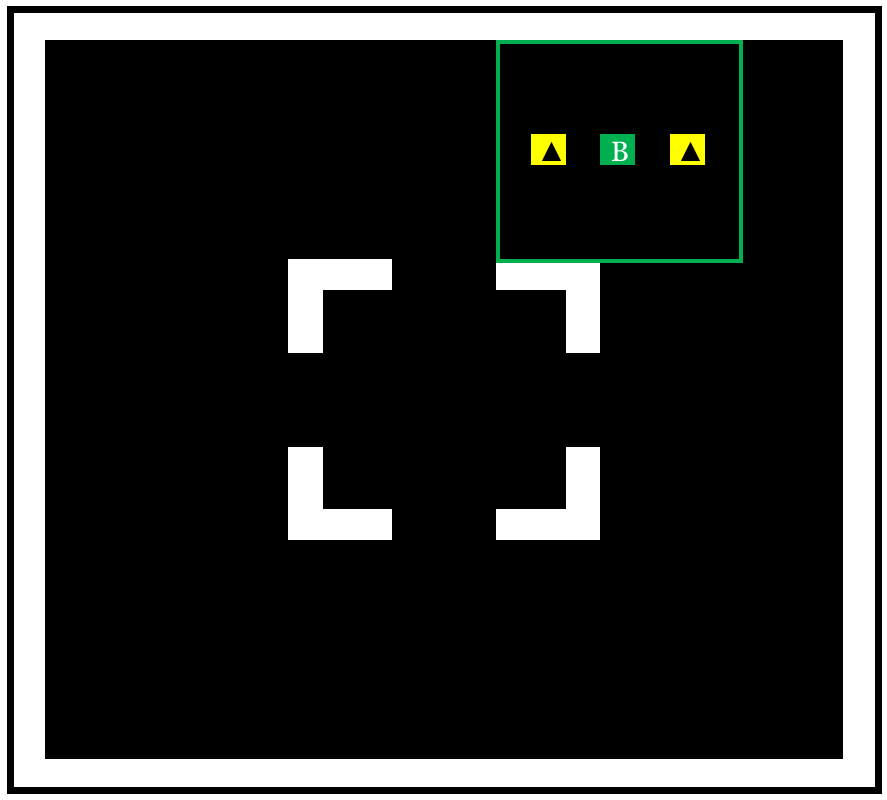}
    \end{minipage}
    \begin{minipage}[t]{0.48\hsize}
      \centering
      \includegraphics[keepaspectratio, width=\linewidth]{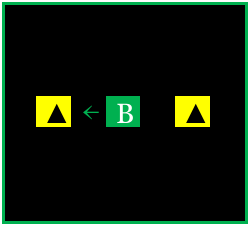}
    \end{minipage}\\
    \begin{minipage}[t]{0.48\hsize}
      \centering
      \includegraphics[keepaspectratio, width=\linewidth]{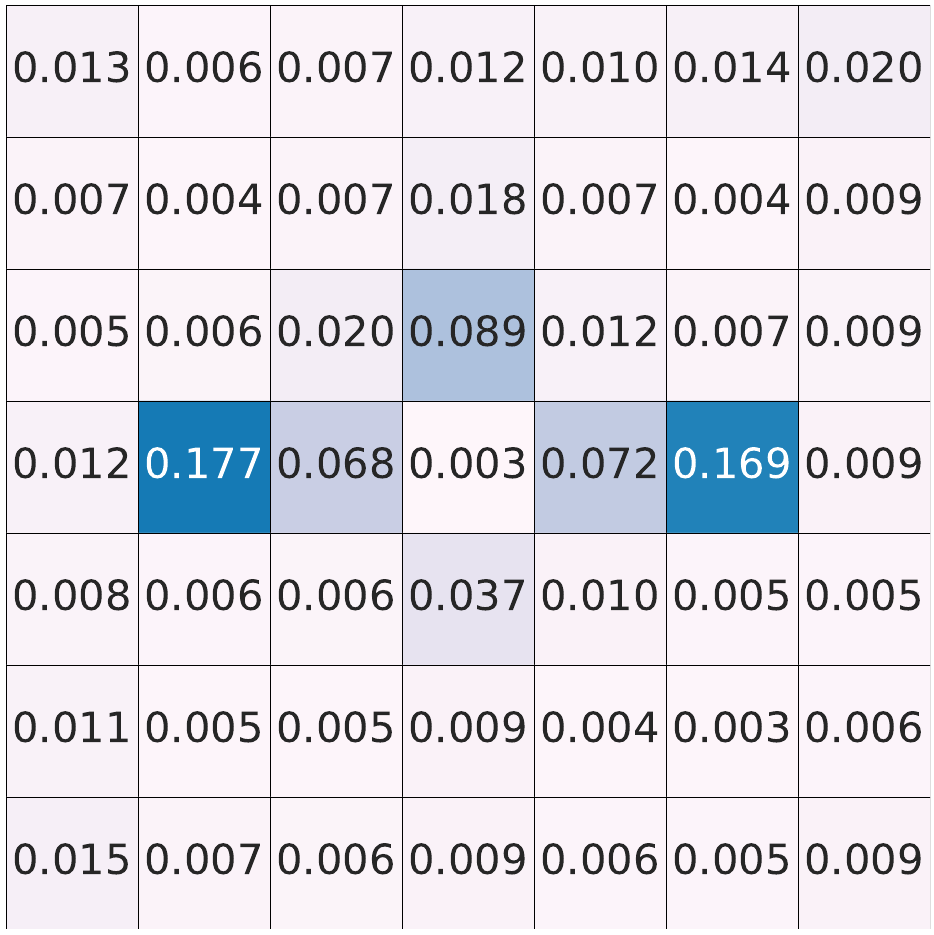}
    \end{minipage}
    \begin{minipage}[t]{0.48\hsize}
      \centering
      \includegraphics[keepaspectratio, width=\linewidth]{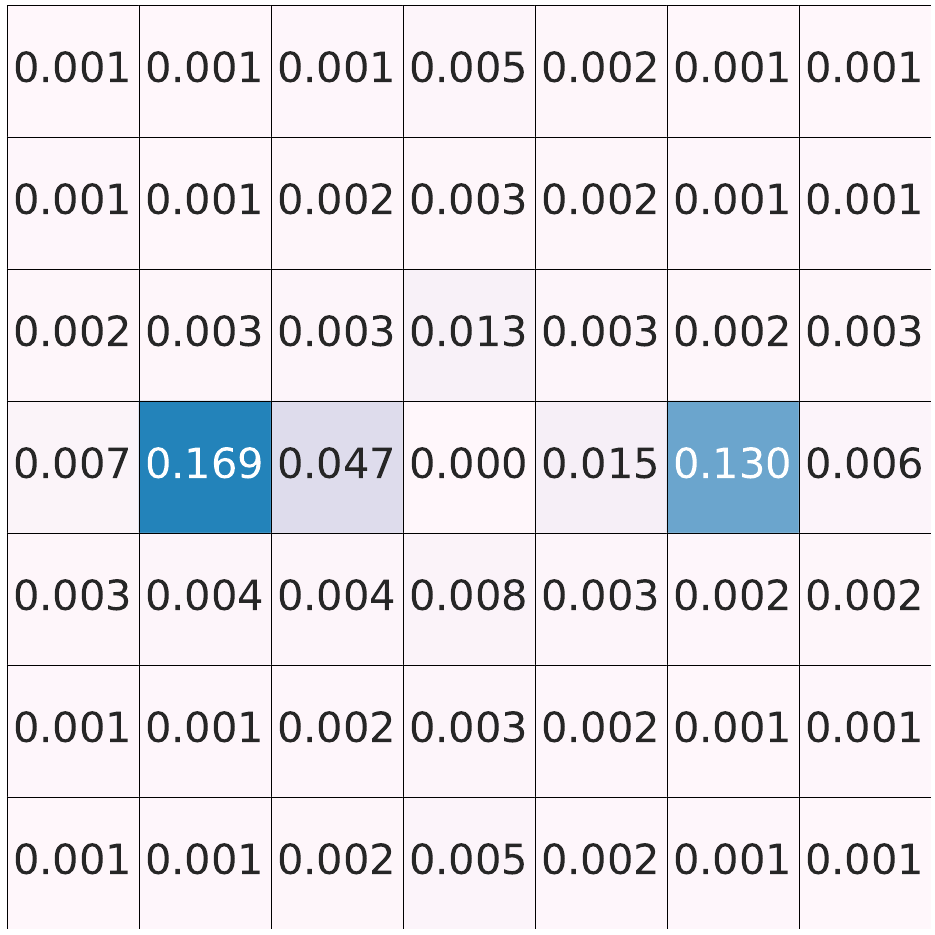}
    \end{minipage}\\
    \subcaption{Solo.}\label{fig:analysis-Opos-A}
  \end{minipage}
  \begin{minipage}[t]{0.48\hsize}
    \centering
    \begin{minipage}[t]{0.48\hsize}
      \centering
      \includegraphics[keepaspectratio, width=\linewidth]{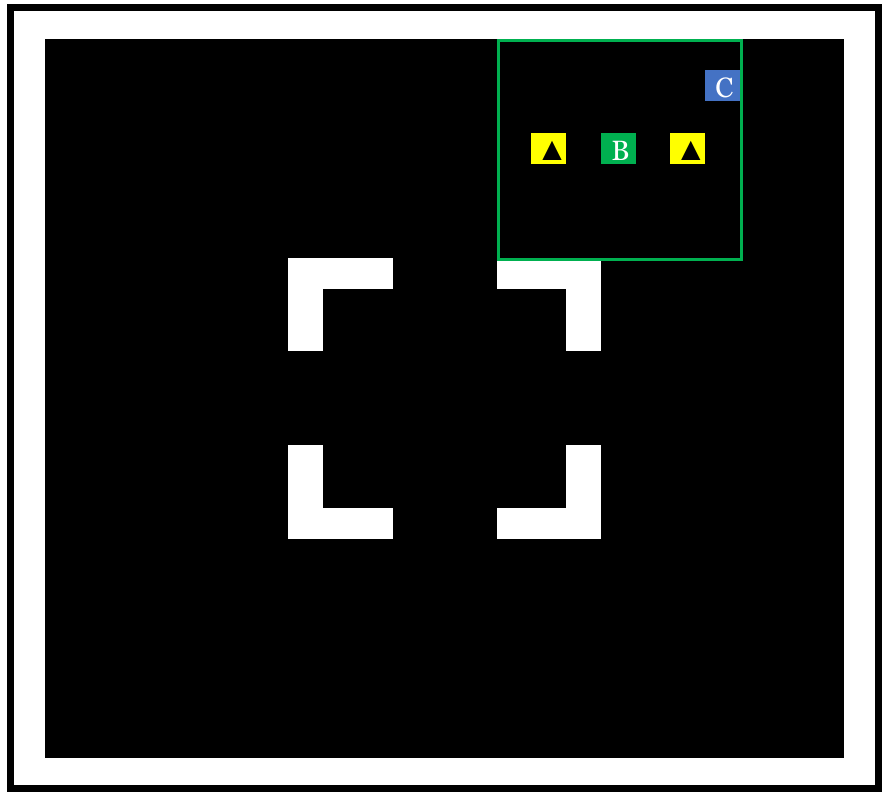}
    \end{minipage}
    \begin{minipage}[t]{0.48\hsize}
      \centering
      \includegraphics[keepaspectratio, width=\linewidth]{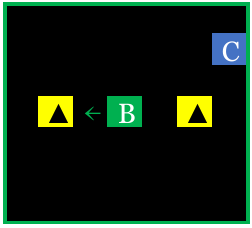}
    \end{minipage}\\
    \begin{minipage}[t]{0.48\hsize}
      \centering
      \includegraphics[keepaspectratio, width=\linewidth]{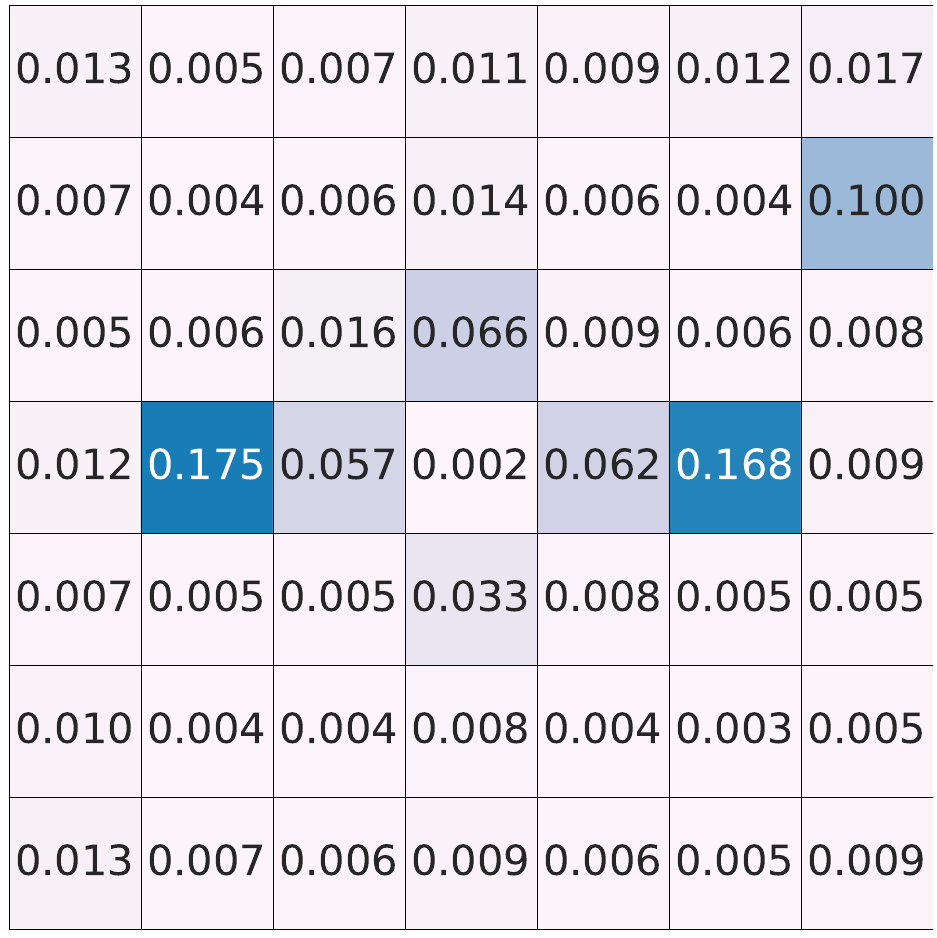}
    \end{minipage}
    \begin{minipage}[t]{0.48\hsize}
      \centering
      \includegraphics[keepaspectratio, width=\linewidth]{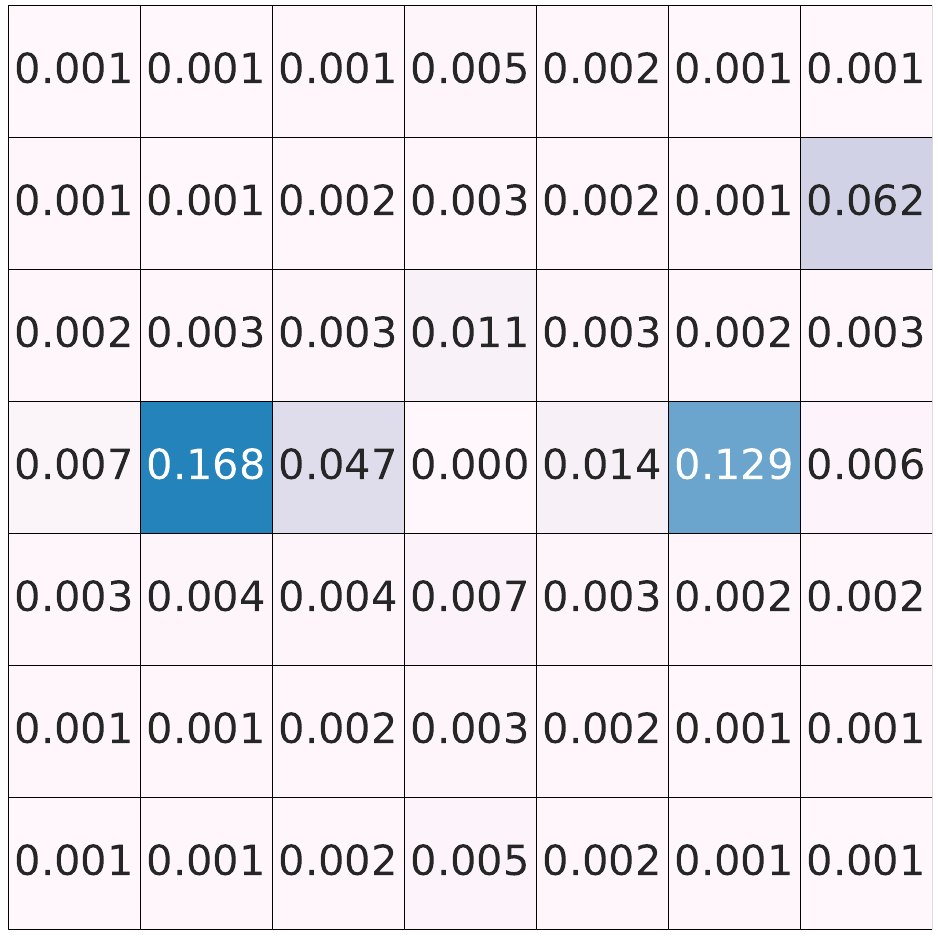}
    \end{minipage}\\
    \subcaption{With \emph{type E} agent.}\label{fig:analysis-Opos-B}
  \end{minipage}
  \vspace{0.5\baselineskip}\\
  \begin{minipage}[t]{0.48\hsize}
    \centering
    \begin{minipage}[t]{0.48\hsize}
      \centering
      \includegraphics[keepaspectratio, width=\linewidth]{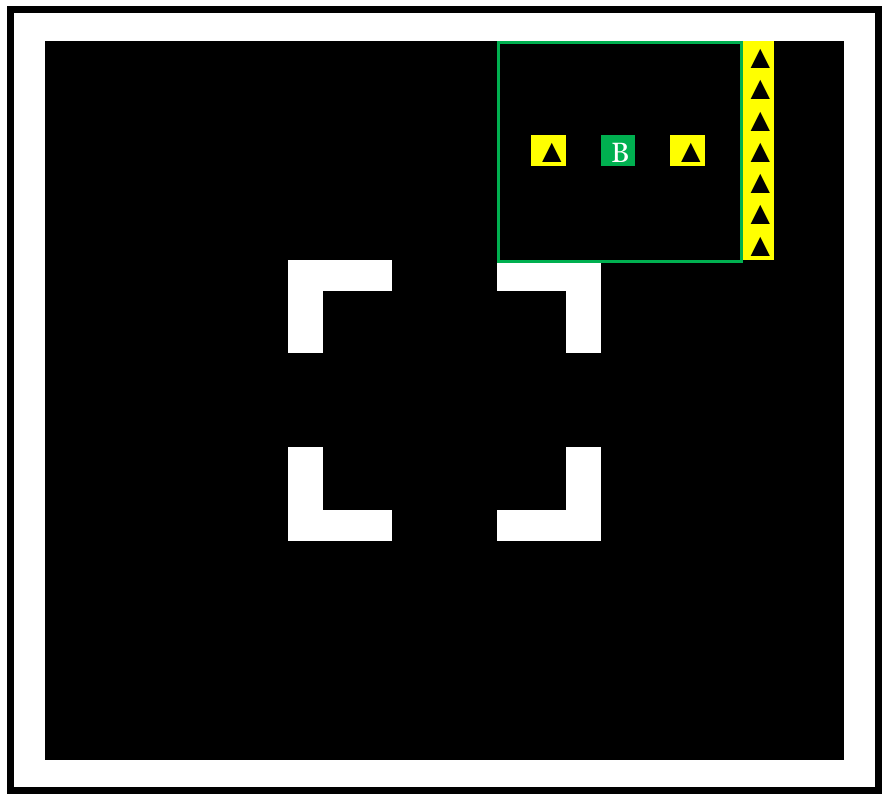}
    \end{minipage}
    \begin{minipage}[t]{0.48\hsize}
      \centering
      \includegraphics[keepaspectratio, width=\linewidth]{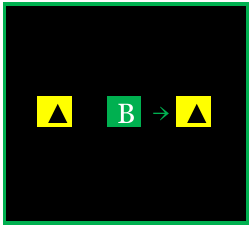}
    \end{minipage}\\
    \begin{minipage}[t]{0.48\hsize}
      \centering
      \includegraphics[keepaspectratio, width=\linewidth]{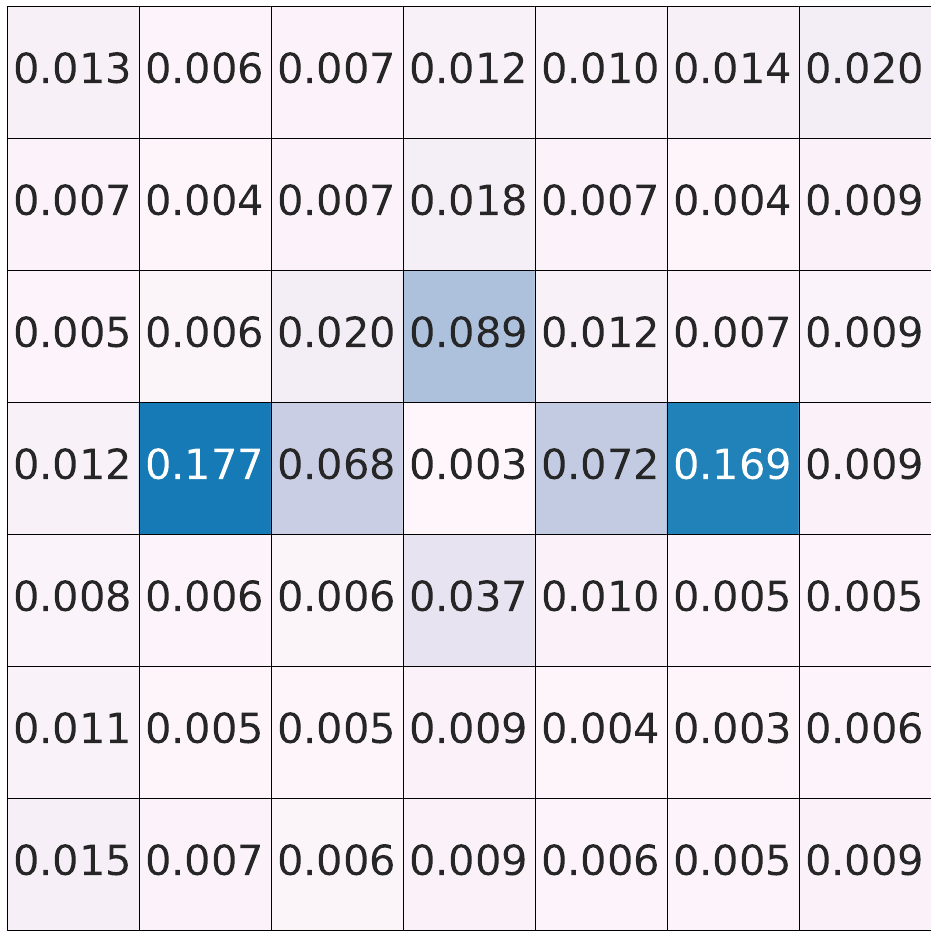}
    \end{minipage}
    \begin{minipage}[t]{0.48\hsize}
      \centering
      \includegraphics[keepaspectratio, width=\linewidth]{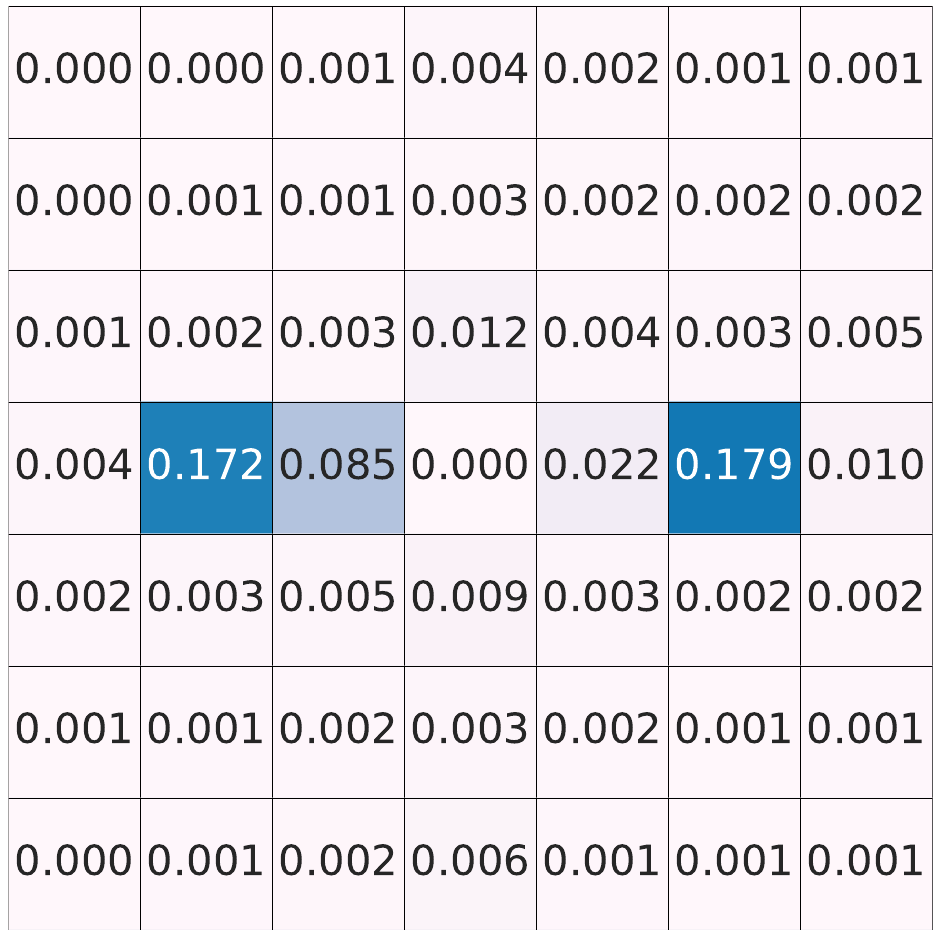}
    \end{minipage}\\
    \subcaption{Solo when many objects exist.}\label{fig:analysis-Opos-C}
  \end{minipage}
  \begin{minipage}[t]{0.48\hsize}
    \centering
    \begin{minipage}[t]{0.48\hsize}
      \centering
      \includegraphics[keepaspectratio, width=\linewidth]{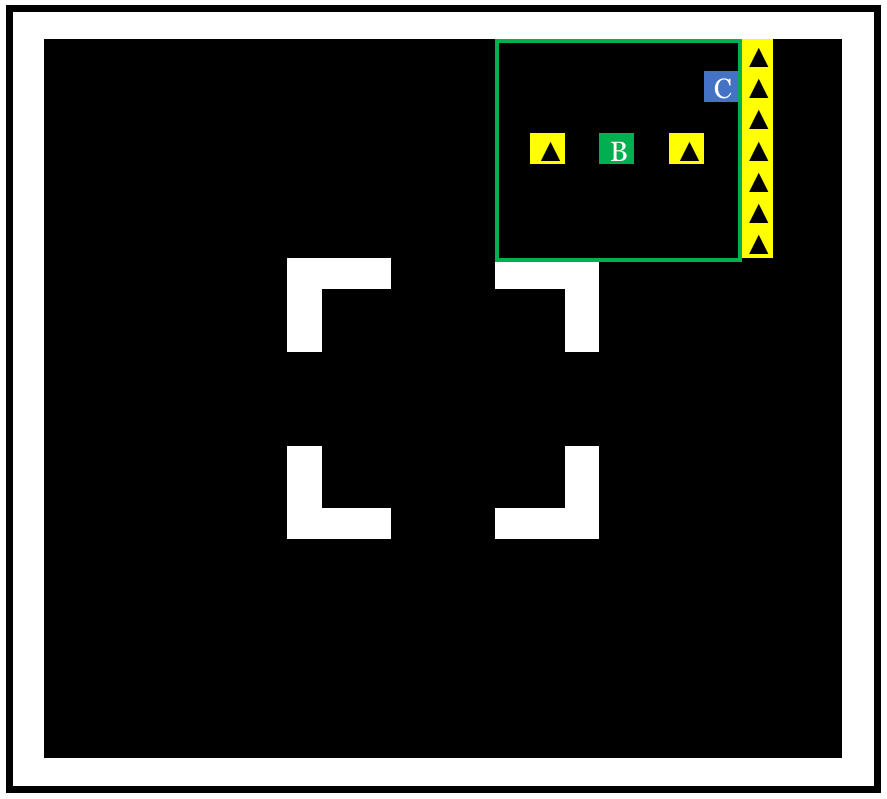}
    \end{minipage}
    \begin{minipage}[t]{0.48\hsize}
      \centering
      \includegraphics[keepaspectratio, width=\linewidth]{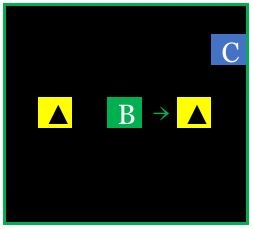}
    \end{minipage}\\
    \begin{minipage}[t]{0.48\hsize}
      \centering
      \includegraphics[keepaspectratio, width=\linewidth]{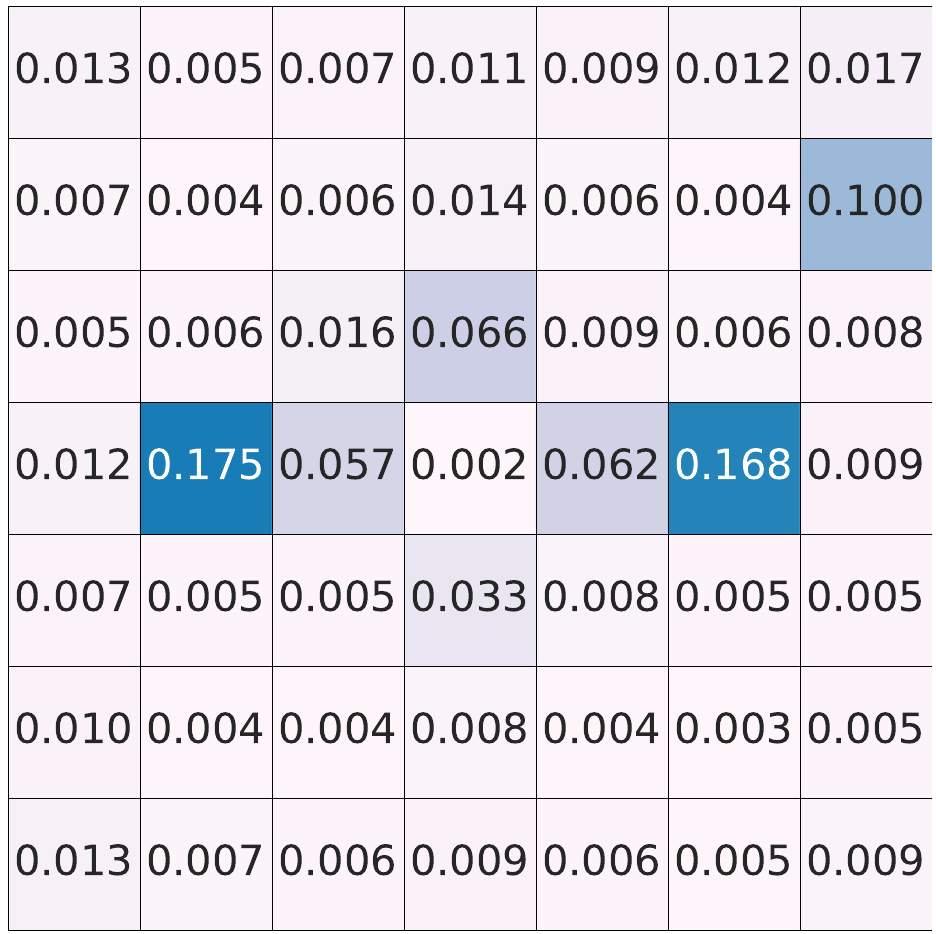}
    \end{minipage}
    \begin{minipage}[t]{0.48\hsize}
      \centering
      \includegraphics[keepaspectratio, width=\linewidth]{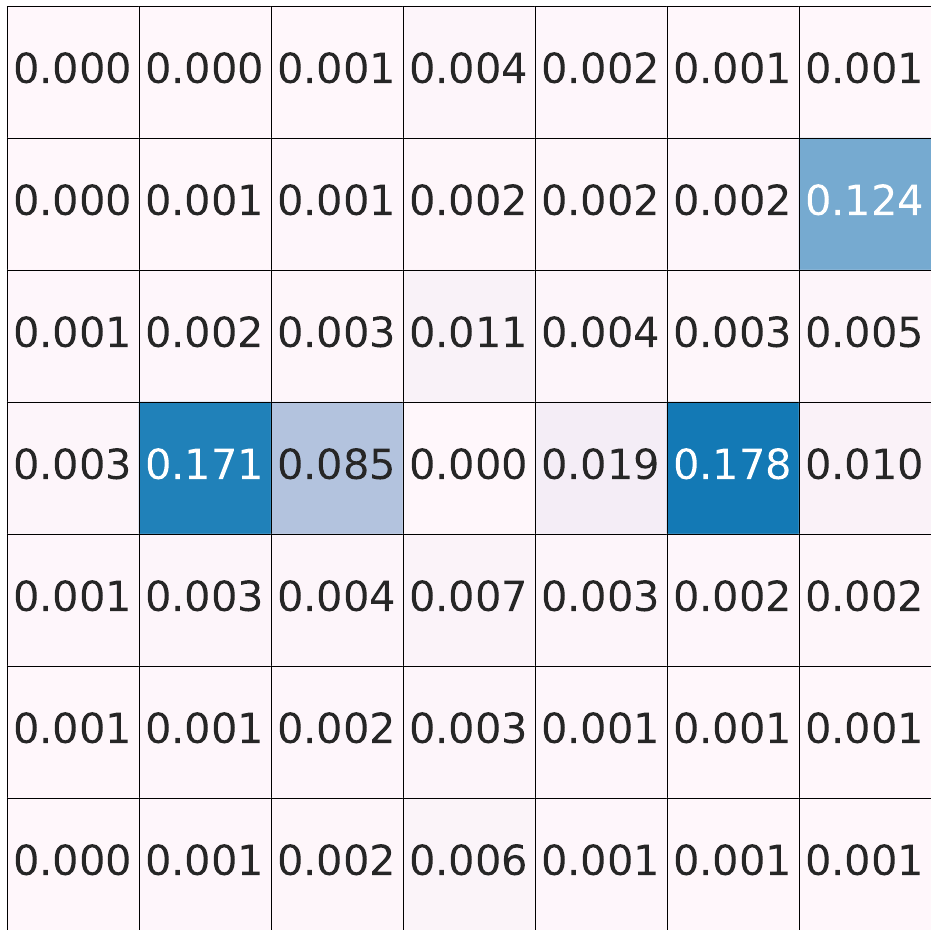}
    \end{minipage}\\
    \subcaption{With opponent and objects.}\label{fig:analysis-Opos-D}
  \end{minipage}
  \caption{Coordination study with \emph{G pos} + \emph{O pos}.}\label{fig:analysis-Opos}
\end{figure}

{\bfseries When \emph{G pos} and \emph{O pos} are available:}
According to Fig.~\ref{fig:performance-result} and Table~\ref{table:quantitative-result}, DA6-X agents achieved even better performance when their global position and objects' positions were given. Similar to the previous cases, the attention heatmaps of DA3-DQN and DA6-DQN are analyzed to compare and understand the improvement in performance. In this study, two $\blacktriangle$ objects were set at the same distance around the \emph{type B} agent, and the behavior was analyzed by placing the \emph{type C} agent close and seven objects outside of the visible range of the \emph{type B} agent, as illustrated in Fig.~\ref{fig:analysis-Opos}. The colors in Fig.~\ref{fig:analysis-Opos} indicate the same roles as those in Fig.~\ref{fig:analysis-Gpos}.
\par

When the \emph{type B} agent observes two objects at the same distance in $\Delta$, it moves toward the object on the left, which has a higher attention weight ($0.177$ of DA3-DQN and $0.169$ of DA6-DQN attention) than that of the object on the right ($0.169$ of DA3-DQN and $0.130$ of DA6-DQN attention), as shown in Fig.~\ref{fig:analysis-Opos-A}.
\par

The same situation was tested when the \emph{type C} agent was located inside the visible range (Fig.~\ref{fig:analysis-Opos-B}). In this case, the \emph{type B} agent accorded relatively less attention ($0.100$ of DA3-DQN and $0.062$ of DA6-DQN attention) to the \emph{type C} agent and moved toward the left object. This behavior is reasonable because the \emph{type B} agent always intended to collect the left object and the existence of the \emph{type C} agent on the right did not affect its policy, as shown in the attention heatmap.
\par

Seven objects were added at four units to the right of the \emph{type B} agent to examine how DA6-IQN agents leverage reusing the saliency vector, which represents \emph{G pos} and \emph{O pos} conditions. Interestingly, the \emph{type B} agent places a high attention weight on the right object ($0.179$ of DA6-DQN attention) and lower attention on the left object ($0.172$ of DA6-DQN attention) after considering the location of seven objects outside of the visible range, as shown in Fig.~\ref{fig:analysis-Opos-C}. The agent then moved right to approach more objects. It is verified that DA6-IQN agents successfully understand the approximate locations of other objects outside their vision from the saliency vector and appropriately build efficient strategies.
\par

Finally, the behavior of the \emph{type B} agent was tested when the \emph{type C} agent was around and seven objects were located on the right. According to the attention heatmap in Fig.~\ref{fig:analysis-Opos-D}, the agent placed high attention weights of $0.178$ and $0.124$ on the right object and \emph{type C} agent, respectively. The \emph{type B} agent assigned almost twice more DA6-DQN attention to the \emph{type C} agent after considering the existence of seven objects outside the visible range because the \emph{type C} agent can be a competitor collecting the same objects. Indeed, the \emph{type B} agent moved toward objects on the right along with the \emph{type C} agent and kept locking on the \emph{type C} agent.
\par

Because DA6-X reuses the saliency vector in CM and local transformer, we can analyze how the conditional states affect the way of obtaining local observation and leads to the coordination. Hence, DA6-X achieves higher interpretability while improving the learning performance, and it can also deal with multiple conditional states, unlike the baseline algorithms.
\par

\section{Conclusion and Discussions}\label{sec:conclusion}
In this study, DA6-X was proposed to improve the interpretability of agents by reusing the saliency vector. Experiments were conducted while providing several conditional states for the object collection game to validate the effectiveness of the proposed method. The result quantitatively demonstrates that DA6-X agents successfully build more efficient policies than baseline methods by reusing the saliency vector, which represents the conditional states of the environment. Analysis of the attention heatmaps generated from the attention weights in the local transformer encoder indicated that providing the conditional states impproves the local observation and coordination strategy of agents and the decision-making process, which had been unknown because of the black-box issue.
\par

This work can be extended to a continuous environment beyond the object collection game. The coordination study of DA6-X agents in a different environment may be in high demand for a better understanding of MADRL and XRL. In addition, the learning performance may improve by replacing transformer encoders in DA6-X with an alternative transformer encoder, such as \emph{TAFA} introduced by \cite{9463791} or \emph{GTrXL}~\cite{parisotto2020stabilizing}.
\par

Our approach may still have limitations in evaluating the interpretability of agents quantitatively. In this work, the mechanism of DA6-X agents observing the environment was introduced via attention heatmaps, qualitatively demonstrating their explainability in several situations. The lack of quantitative evaluation on the explainability may be fatal in critical applications, such as self-driving systems, as in these systems high safety and algorithmic accountability need to be ensured. Therefore, our next research step is to quantitatively examine the transparency of agents' decision-making process via attention heatmaps.
\par

\section*{Acknowledgment}
This work was partly supported by JSPS KAKENHI Grant Numbers 20H04245.

% \bibliographystyle{plain}
% \bibliography{reference}

\end{document}